\documentclass[10pt,twocolumn,letterpaper]{article}

\usepackage{cvpr}              % To produce the CAMERA-READY version
\usepackage{multirow}
\usepackage[table,xcdraw]{xcolor}
\usepackage{adjustbox}  % 比 \resizebox 更灵活
\usepackage{makecell}   % 可选：让表头断行更容易

\usepackage{amsmath}     % 数学符号和字体
\usepackage{amssymb}     % 额外的数学符号
\definecolor{highlight}{RGB}{204,102,0} % 橙色
\usepackage{stfloats}

\usepackage{tabularx}
\usepackage{float}

\definecolor{cvprblue}{rgb}{0.21,0.49,0.74}
\usepackage[pagebackref,breaklinks,colorlinks,allcolors=cvprblue]{hyperref}

\def\paperID{39247} % *** Enter the Paper ID here
\def\confName{CVPR}
\def\confYear{2026}

\title{Focus-to-Perceive Representation Learning: A Cognition-Inspired Hierarchical Framework for Endoscopic Video Analysis}

\author{Yuan Zhang$^{1,2}$, Sihao Dou$^{1,2}$, Kai Hu$^{1,2,*}$, Shuhua Deng$^{1,2}$, Chunhong Cao$^{1,2}$, Fen Xiao$^{1,2}$, Xieping Gao$^{3}$\\
\small $^{1}$Key Laboratory of Intelligent Computing and Information Processing of Ministry of Education, Xiangtan University\\
\small $^{2}$School of Computer Science, Xiangtan University\\
\small $^{3}$Key Laboratory for Artificial Intelligence and International Communication, Hunan Normal University\\
\tt\small \{yuanz, kaihu, shuhuadeng, caoch, xiaof\}@xtu.edu.cn\\
\tt\small shdou@smail.xtu.edu.cn \hspace{5em} xpgao@hunnu.edu.cn }

\begin{document}
\maketitle
\begingroup
\renewcommand\thefootnote{\fnsymbol{footnote}}
\setcounter{footnote}{1} % * 
\footnotetext{Corresponding author.}
\endgroup

\begin{abstract}

% Endoscopic video analysis is crucial for early gastrointestinal screening, but its progress is constrained by limited high-quality annotations. While self-supervised video pre-training shows promise, existing methods designed for natural videos tend to prioritize dense spatio-temporal modeling and exhibit motion bias, neglecting the static, structured semantics that are critical for clinical decision-making.
Endoscopic video analysis is essential for early gastrointestinal screening but remains hindered by limited high-quality annotations. While self-supervised video pre-training shows promise, existing methods developed for natural videos prioritize dense spatio-temporal modeling and exhibit motion bias, overlooking the static, structured semantics critical to clinical decision-making.
% To address this challenge, we propose \textbf{F}ocus-to-\textbf{P}erceive \textbf{R}epresentation \textbf{L}earning (\textbf{FPRL}), a cognition-inspired hierarchical framework that emulates the clinical examination process of endoscopic videos. FPRL first focuses on intra-frame lesion-centric regions to learn static semantics, and then perceives their evolution across frames to model contextual semantics. To achieve this, FPRL employs a hierarchical semantic modeling mechanism that explicitly distinguishes and collaboratively learns both types of semantics. Specifically, it begins by capturing static semantics through the application of teacher-prior adaptive masking (TPAM) combined with multi-view sparse sampling. This approach mitigates redundant temporal dependencies and enables the model to concentrate on lesion-related local semantics. Following this, contextual semantics are derived through cross-view masked feature completion (CVMFC) and attention-guided temporal prediction (AGTP). These processes establish cross-view correspondences and effectively model structured inter-frame evolution, thereby reinforcing temporal semantic continuity while preserving global contextual integrity. Extensive experiments on 11 endoscopic video datasets show that FPRL achieves state-of-the-art performance across diverse downstream tasks, demonstrating its effectiveness and strong generalization in endoscopic video representation learning.
To address this challenge, we propose \textbf{F}ocus-to-\textbf{P}erceive \textbf{R}epresentation \textbf{L}earning (\textbf{FPRL}), a cognition-inspired hierarchical framework that emulates clinical examination. FPRL first focuses on intra-frame lesion-centric regions to learn static semantics, and then perceives their evolution across frames to model contextual semantics. To achieve this, FPRL employs a hierarchical semantic modeling mechanism that explicitly distinguishes and collaboratively learns both types of semantics. Specifically, it begins by capturing static semantics via teacher-prior adaptive masking (TPAM) combined with multi-view sparse sampling. This approach mitigates redundant temporal dependencies and enables the model to concentrate on lesion-related local semantics. Following this, contextual semantics are derived through cross-view masked feature completion (CVMFC) and attention-guided temporal prediction (AGTP). These processes establish cross-view correspondences and effectively model structured inter-frame evolution, thereby reinforcing temporal semantic continuity while preserving global contextual integrity. Extensive experiments on 11 endoscopic video datasets show that FPRL achieves superior performance across diverse downstream tasks, demonstrating its effectiveness in endoscopic video representation learning.
%For Camera Ready
The code is available at \textcolor{magenta}{https://github.com/MLMIP/FPRL}.

\end{abstract}

\section{Introduction}
\label{sec:introduction}

\begin{figure}[t]
    \centering
    \includegraphics[width=1\linewidth]{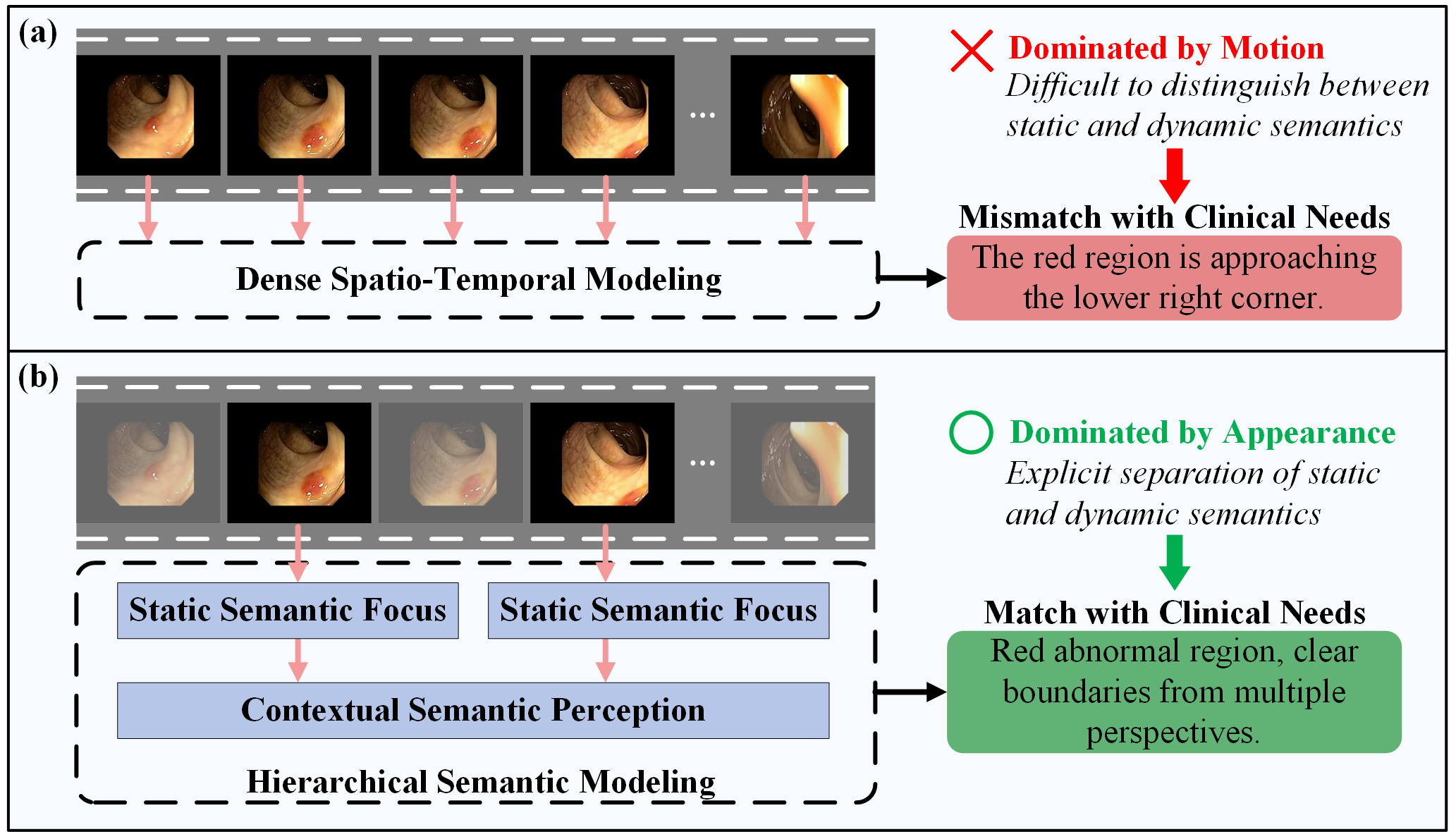}
    \caption{(a) Dense spatio-temporal modeling often captures irrelevant semantics due to motion reliance. (b) Our hierarchical approach isolates static from dynamic context, extracting appearance features required for diagnosis.}
    \label{fig:introfig}
\end{figure}

Endoscopic video analysis plays a crucial role in the early screening and diagnosis of gastrointestinal diseases by facilitating more accurate lesion detection and assisting clinicians in making timely decisions. However, its further advancement is significantly hindered by the scarcity of high-quality annotated data, which limits the performance and generalization capabilities of current algorithms. 
To address this challenge, a promising strategy involves leveraging advancements in general video understanding to learn generalizable representations from large-scale unlabeled videos through self-supervised video pre-training \cite{tong2022videomae,bertasius2021spaceTimesformer,wang2023videomaev2}.
In these methods, carefully designed pretext tasks (\textit{e.g.}, frame reconstruction, temporal prediction, or masked completion) are employed to capture spatio-temporal consistency without relying on human annotations, thereby substantially reducing dependence on limited high-quality labels.

% At present, most of the existing self-supervised video pre-training methods are developed on natural videos, and their core idea is to learn semantics through dense spatio-temporal modeling, which has proven highly effective for tasks such as action recognition 
Most existing self-supervised video pre-training methods are developed on natural videos, where dense spatio-temporal modeling effectively captures motion semantics for tasks such as action recognition \cite{tong2022videomae,bertasius2021spaceTimesformer,wang2023videomaev2}. 
% However, there are fundamental differences between natural and endoscopic videos in terms of their semantic structures. Natural videos heavily rely on prominent temporal dynamics to capture motion semantics and contextual evolution. In contrast, endoscopic videos primarily derive semantics from static, localized, and highly structured visual cues \cite{ji2022videopolyp}. 
However, endoscopic videos differ fundamentally from natural videos in semantic structure, as they rely primarily on static, localized visual cues \cite{ji2022videopolyp} rather than prominent temporal dynamics.
Therefore, when dense spatio-temporal paradigms are directly applied to endoscopic videos \cite{hu2024mmcrl,tian2025endomamba,wang2025improvingEndoFM-LV,wang2023foundationEndoFM,ali2021deependoscopy,ji2022videopolyp}, an inherent contradiction arises: models tend to focus on non-semantic motions such as camera jitter and tissue displacement while underutilizing critical static semantics like the morphology, color, and texture of the lesion \cite{ji2022videopolyp,jinhao2024PSTUDA}.
We refer to this systematic tendency to overemphasize task-independent dynamic patterns while underestimating static cues of diagnostic significance as “motion bias”, which ultimately weakens semantic focus and thereby reduces cross-task generalization ability.

Although this motion bias can be partially mitigated through strategies such as sparse sampling \cite{gupta2023siameseMAE,pei2024videomac,liu2025futureTCoRe}, temporal masking \cite{tong2022videomae,wang2023videomaev2}, and local attention mechanisms \cite{bertasius2021spaceTimesformer}, it is important to note that these techniques were originally developed for natural video tasks like action recognition. Their primary objective was to enhance the efficiency and robustness of temporal modeling rather than explicitly addressing motion bias. As a result, there remains a significant gap in the explicit modeling paradigms specifically designed to accommodate the static, localized, and structured semantic characteristics prevalent in endoscopic videos, warranting further investigation.

In clinical practice, experienced endoscopists do not primarily rely on motion to interpret endoscopic videos. Instead, they typically begin by meticulously examining individual frames to identify semantically salient regions, such as subtle anomalies in color, texture, or lesion boundaries. This initial analysis is followed by tracing the temporal evolution of these candidate regions to infer pathological characteristics \cite{hirsch2023selfendoscopic}. This process can be summarized as first focusing on diagnostically meaningful static cues and then perceiving their changes over time while largely disregarding irrelevant motions such as camera jitter or tissue displacement. 
% Inspired by the cognitive pattern of “focus before perception”, we propose \textbf{F}ocus-to-\textbf{P}erceive \textbf{R}epresentation \textbf{L}earning (\textbf{\textit{FPRL}}), a novel framework for endoscopic video analysis. The \textit{FPRL} framework employs a semantics-driven hierarchical modeling mechanism that enhances spatio-temporal semantic coordination, effectively mitigating motion bias and improving representation learning (see Fig.~\ref{fig:introfig}).
Inspired by the “focus before perception” cognitive pattern, we propose \textbf{F}ocus-to-\textbf{P}erceive \textbf{R}epresentation \textbf{L}earning (\textbf{\textit{FPRL}}), a novel semantics-driven hierarchical framework that enhances spatio-temporal semantic coordination to mitigate motion bias and improve representation learning for endoscopic video analysis (see Fig.~\ref{fig:introfig}).

% Specifically, our proposed \textit{FPRL} framework performs hierarchical semantic modeling by explicitly disentangling and jointly learning both static and contextual semantics. 
Specifically, \textit{FPRL} performs hierarchical semantic modeling by explicitly disentangling and jointly learning both static and contextual semantics. 
% First, we introduce a teacher-prior adaptive masking (TPAM) strategy combined with multi-view sparse sampling to suppress redundant temporal dependencies and accentuate lesion-related semantic saliency. This process effectively counteracts low-level dynamic noise and non-semantic motion, yielding a robust static representation. 
First, a teacher-prior adaptive masking (TPAM) strategy combined with multi-view sparse sampling suppresses redundant temporal dependencies and accentuates lesion-related semantic saliency. This process effectively counteracts low-level dynamic noise and non-semantic motion, yielding a robust static representation. 
% Subsequently, we employ the cross-view masked feature completion (CVMFC) and attention-guided temporal prediction (AGTP) to model the temporal evolution and structural correlation of semantic regions. This process establishes consistent inter-frame semantics and preserves global contextual integrity, thereby capturing coherent dynamic patterns. 
Subsequently, cross-view masked feature completion (CVMFC) and attention-guided temporal prediction (AGTP) model the temporal evolution and structural correlation of semantic regions. This process establishes consistent inter-frame semantics and preserves global contextual integrity, thereby capturing coherent dynamic patterns. 
Through this semantics-driven hierarchical approach, \textit{FPRL} enhances semantic discriminability while retaining temporal coherence, thus substantially alleviating motion bias and improving generalization across downstream tasks. The framework establishes an innovative, cognition-inspired paradigm for endoscopic video representation learning.

In summary, our main contributions are as follows:
\begin{itemize}
    \item We are the first to develop \textit{FPRL}, a cognition-inspired representation learning framework that employs a hierarchical semantic modeling mechanism to separate and jointly learn static and contextual semantics, thereby mitigating motion bias and achieving robust representations for endoscopic video analysis.
    \item We propose a \textit{teacher-prior adaptive masking} strategy to emphasize lesion-centric static semantics, and combine it with \textit{cross-view masked feature completion} and \textit{attention-guided temporal prediction} to ensure temporal coherence and global integrity of contextual semantics.
    \item Extensive experiments conducted on 11 endoscopic video datasets show that \textit{FPRL} consistently outperforms existing methods across diverse downstream tasks including classification, segmentation, detection, and recognition, demonstrating its effectiveness and strong generalization in endoscopic video representation learning.
\end{itemize}

\section{Related Work}
\label{sec:related}

\subsection{Self-supervised Video Representation Learning}

Self-supervised learning (SSL) \cite{yuzhang2025confusiondriven,he2022MAE,grill2020BYOL} utilizes the intrinsic structure of data to generate supervisory signals, thereby facilitating the acquisition of transferable feature representations without the need for manual annotation. The predominant approaches in self-supervised video learning encompass discriminative methods that leverage temporal consistency or contrastive learning \cite{jenni2021timeCorrespendence1,qian2021timeCorrespendence2,sermanet2018timeCorrespendence3,qing2023hierarchicalconsistency}, reconstruction-based masked modeling techniques such as MVM/MFM \cite{tong2022videomae,wang2023videomaev2,liu2025futureTCoRe}, and the emerging joint embedding predictive architecture paradigm \cite{bardes2024vjepa}, which conducts predictions directly within the representation space.

Early self-supervised video methods mainly transferred image-based pretext tasks to video sequences, such as predicting frame order \cite{lee2017shuffleFrame2,misra2016shuffleFrame}, playback speed \cite{benaim2020speednet}, or future frames \cite{srivastava2015PredictFuture} to capture temporal dependencies, assuming motion conveys semantic information. With the rise of the contrastive learning paradigm \cite{caron2021DINO,oquab2023dinov2,chen2020SimCLR,grill2020BYOL}, the research focus shifted to leveraging inter-sample consistency to learn spatio-temporal representations \cite{jenni2021timeCorrespendence1,qian2021timeCorrespendence2,sermanet2018timeCorrespendence3}. These techniques construct positive and negative pairs using temporal continuity, semantic similarity, or motion cues, achieving notable advances in downstream tasks such as action recognition \cite{actionrec2021large-scale,kuehne2011motionrec} and video retrieval \cite{liu2024retrieval,kordopatis2019retrieval}. 
Alternatively, methods inspired by masked autoencoders \cite{he2022MAE} learn visual co-occurrence patterns by reconstructing randomly masked portions of the input. Representative approaches like VideoMAE \cite{tong2022videomae}, VideoMAE V2 \cite{wang2023videomaev2}, and ST-MAE \cite{feichtenhofer2022ST-MAE} successfully extend masked image modeling to video with various spatio-temporal masking strategies. To further address high video redundancy, subsequent works have explored motion-guided adaptive masking \cite{fan2023motionMGMAE,bandara2023adamae}, sparse sampling \cite{pei2024videomac,gupta2023siameseMAE,hernandez2024vicmae,liu2025futureTCoRe}, and latent-space reconstruction \cite{bardes2024vjepa} for improved semantic emphasis and training efficiency.

% Overall, these approaches share a fundamental assumption that the semantics of video content are encoded within temporal dynamics. This perspective renders dense spatiotemporal modeling adequate for capturing consistency in natural videos, where the semantics of actions correlate with variations in motion.
Overall, these methods share a fundamental assumption that video semantics are encoded within temporal dynamics. This perspective renders dense spatiotemporal modeling adequate for capturing consistency in natural videos, where action semantics correlate with variations in motion.

\subsection{Self-supervised Learning for Endoscopic Video}

In endoscopic video analysis, SSL addresses the challenge of annotation scarcity by learning from unlabeled data. Early approaches employed contrastive or pretext tasks \cite{hirsch2023selfendoscopic,vats2021endossl3,intrator2023endossl,mohammed2020endossl2} with spatio-temporal augmentations, achieving initial improvements in recognition and retrieval tasks.
Recent approaches have adapted innovations from natural video SSL. EndoFM \cite{wang2023foundationEndoFM} and EndoFM-LV \cite{wang2025improvingEndoFM-LV} implement contrastive learning across spatio-temporal views, with the latter extending to minute-level sequences to enhance temporal modeling capabilities. Concurrently, M$^2$CRL \cite{hu2024mmcrl} introduces masked modeling techniques that utilize both spatial and temporal masking to improve contextual reconstruction. Furthermore, EndoMamba \cite{tian2025endomamba} delves into long-sequence modeling through the Mamba architecture \cite{gu2024mamba,gu2022S4,gu2020hippo,liu2024vmamba,zhu2024visionmamba}, incorporating hierarchical pretraining to differentiate between high-level and low-level semantics.

Nevertheless, these methods predominantly adhere to a motion-centric paradigm, which poses significant challenges for endoscopic analysis where diagnostic semantics are primarily dependent on static morphological features such as the color and texture of the lesion \cite{ji2022videopolyp}. As a result, motion-focused objectives often overfit non-semantic artifacts (\textit{e.g.}, camera jitter or tissue displacement), while inadequately representing critical static features essential for accurate diagnosis \cite{ali2021deependoscopy,ji2022videopolyp}.

\subsection{Hierarchical Representation Learning}

Recent advancements in self-supervised video learning have increasingly emphasized hierarchical representation learning \cite{xiao2022hierarchicalsep,qing2023hierarchicalconsistency}. This approach effectively disentangles complementary semantic cues by organizing video content into multi-level abstractions, typically distinguishing static spatial semantics from dynamic temporal contexts, to yield structured and interpretable representations.

In general video domains, hierarchical modeling has been investigated through various methods including spatio-temporal disentanglement \cite{wang2015stsep,diba2018stsep2,qing2023stsep3,korbar2019stsep4}, multi-stage or cross-scale encoding \cite{liu2022multisep2,arnab2021vivitmultisep}, and masked hierarchical reconstruction \cite{feichtenhofer2022ST-MAE,wang2023videomaev2,xiao2022hierarchicalsep,reed2023scalesep}. These strategies are adept at capturing both fine-grained appearance details and long-range temporal dependencies, thereby enhancing generalization and efficiency compared to flat models. However, such hierarchical approaches remain largely underexplored in the context of endoscopic video analysis. Existing methodologies continue to model spatial and temporal information jointly. This overlooks a key endoscopic characteristic that diagnostic semantics arise from structured spatial patterns and their temporal evolution, which are inadequately captured by non-hierarchical modeling.

\section{Methodology}
\label{sec:method}

\begin{figure*}
    \centering
    \includegraphics[width=1\linewidth]{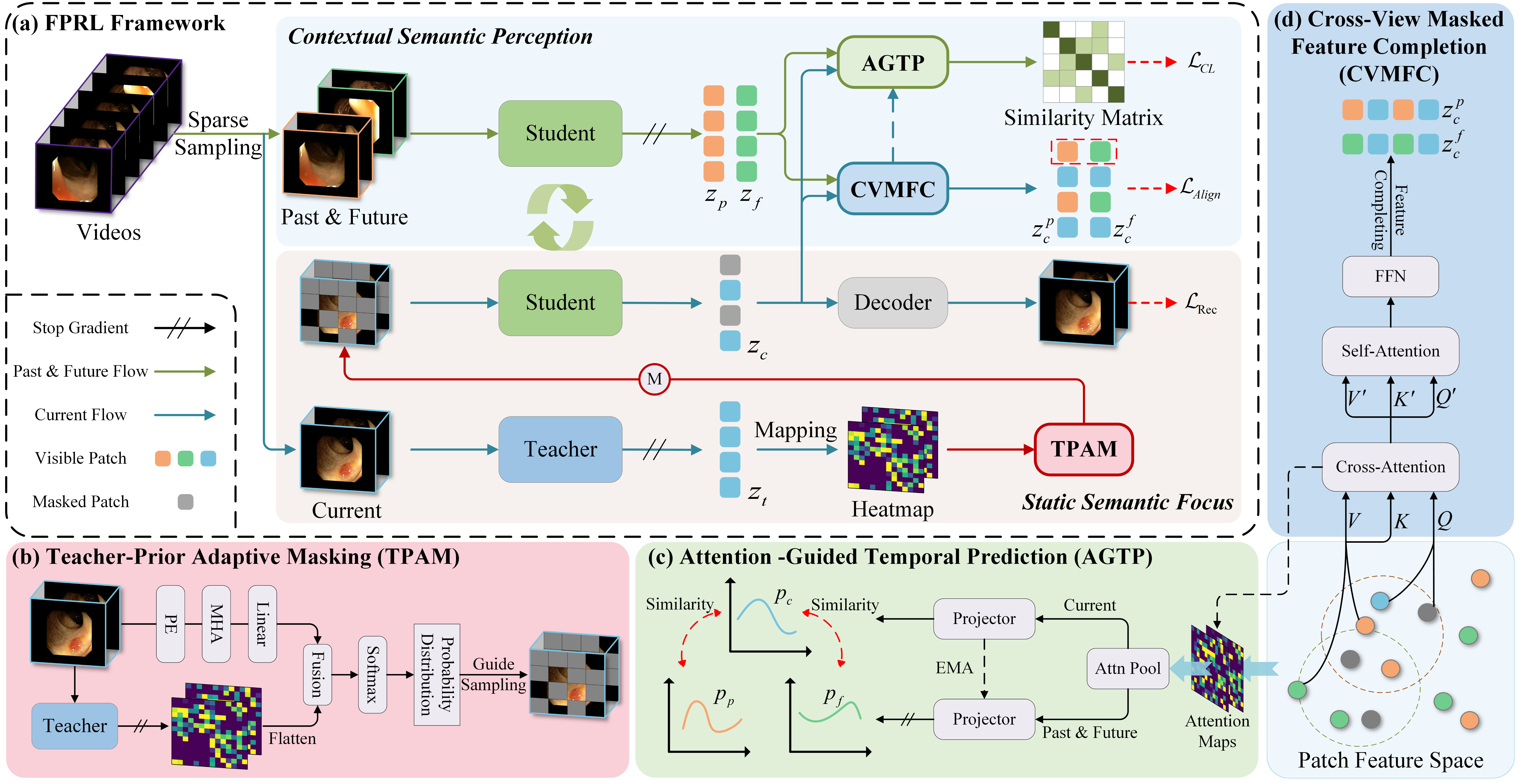}
    \caption{The pipeline of the proposed \textit{FPRL}. For static semantic focus, we employ a teacher-prior adaptive masking strategy coupled with pixel reconstruction. For contextual semantic perception, we use cross-view masked feature completion with latent spatial alignment and an attention-guided temporal prediction module with contrastive learning.}
    \label{fig:FPRL}
\end{figure*}

\subsection{Overview}

As illustrated in Fig. \ref{fig:FPRL}, our proposed \textit{FPRL} is a cognition-inspired hierarchical framework that emulates the clinical examination process of endoscopic videos. 
The aim is first to focus on lesion-centric salient representations and then to enforce cross-view consistency via feature correlation. 
In order to operationalize this cognitive process, we propose a hierarchical semantic modeling mechanism, which consists of two components: \textit{Static Semantic Focus} and \textit{Contextual Semantic Perception}. 
The former is used to capture lesion-centered intra-view semantics, and the latter is employed to model cross-view temporal evolution to maintain contextual consistency. 
This hierarchy refines semantics and explicitly mitigates the motion bias common in endoscopic videos.

To suppress low-level dynamic noise and non-semantic motion, we develop the Teacher-Prior Adaptive Masking (TPAM) based on multi-view sparse sampling in Section \ref{3.2}, which uses teacher-derived saliency features to guide student to focus on lesion-centered local cues to produce robust static representations. 
We then enhance inter-frame semantic consistency and global contextual integrity by the Cross-View Masked Feature Completion (CVMFC) and Attention-Guided Temporal Prediction (AGTP) in Section \ref{3.3}. 
Finally, we optimize a unified objective in Section \ref{3.4} that combines pixel-wise reconstruction, feature alignment, and temporal prediction loss to enable collaborative learning of static and contextual semantics and produce temporally consistent representations.

\subsection{Semantic Focus for Static Representation}
\label{3.2}

To obtain robust intra-view semantics in the presence of non-semantic motion, we propose \textit{Static Semantic Focus}, which concentrates modeling capacity on lesion-centric cues while suppressing redundant temporal correlations introduced by camera jitter and tissue motion. This design produces stable static representations that serve as a reliable anchor for subsequent cross-view modeling.

% -------------------------------------------------------------
% \paragraph{Multi-view Sparse Sampling.}
\noindent\textbf{Multi-view Sparse Sampling.} Given a video sequence 
\(
V = \{F_t \in \mathbb{R}^{H \times W \times C}\}_{t=1}^{T},
\)
we randomly select a temporal window 
\(
\mathcal{W} = [t_a, t_b]
\)
and independently sample 3 sparse views (past, current, and future) without replacement:
\begin{equation}
\mathcal{V}_{\text{p}},~ \mathcal{V}_{\text{c}},~ \mathcal{V}_{\text{f}} \subset \mathcal{W},
\quad 
|\mathcal{V}_{\text{p}}| = |\mathcal{V}_{\text{c}}| = |\mathcal{V}_{\text{f}}| = 2,
\end{equation}

This sampling strategy intuitively suppresses the dynamic redundancy of video input while maintaining semantic diversity across views.
Each view $\mathcal{V}$ is partitioned into non-overlapping patches and linearly embedded into a token sequence:
\begin{equation}
X = \mathrm{Embed}(\mathcal{V}) \in \mathbb{R}^{N \times d},
\end{equation}
where \(N\) denotes the total number of tokens and \(d\) is the embedding dimension.  
We adopt a teacher–student design with the current view processed by the teacher encoder $f_t$:
\begin{equation}
z_t = f_t(X_{\!c}) \in \mathbb{R}^{N \times d},
\end{equation}
while the past and future views are encoded by the student encoder $f_s$ into $z_p$ and $z_f$, respectively. The teacher provides saliency priors and alignment targets (no gradient), and the student is trainable.

% -------------------------------------------------------------
% \paragraph{Semantic Information Extraction.}
\noindent\textbf{Semantic Information Extraction.} We build encoders upon Mamba-based blocks derived from the Selective State Space Model (SSM) \cite{gu2020hippo,gu2024mamba,gu2022S4}, which models a 1D continuous-time input $x(t)$ via a hidden state $h(t)$:
\begin{equation}
h'(t) = A h(t) + B x(t), \quad y(t) = C h(t),
\label{eq:ssm_ode}
\end{equation}
where $A \in \mathbb{R}^{r\times r}$, $B \in \mathbb{R}^{r\times 1}$, and $C \in \mathbb{R}^{1\times r}$. 
Discretization with a learnable timescale $\Delta$ leads to:
\begin{equation}
\left\{
\begin{aligned}
\overline{A} &= \exp(\Delta A),\\
\overline{B} &= (\Delta A)^{-1}\big(\exp(\Delta A) - I\big)\cdot \Delta B,
\end{aligned}
\right.
\label{eq:mamba_discrete}
\end{equation}

The hidden state is then updated recursively as 
\(h_t = \overline{A} h_{t-1} + \overline{B} x_t\), 
allowing long-range dependency modeling through state propagation while maintaining linear-time complexity.

Based on this formulation, we instantiate the student encoder $f_s$ with EndoMamba \cite{tian2025endomamba} trained from scratch, and the teacher encoder $f_t$ with a VideoMamba \cite{li2024videomamba} model pre-trained on large-scale video data (frozen during training). EndoMamba adopts a hybrid topology:
\begin{itemize}
    \item \textbf{Bidirectional Mamba} performs forward and backward state propagation within each frame’s patch sequence to capture intra-frame spatial consistency and detailed texture semantics.
    % \item \textbf{Unidirectional Mamba} applies a single forward state update over all patches across frames to integrate inter-frame contextual dynamics in a temporally causal manner.
    \item \textbf{Unidirectional Mamba} applies a single forward state update over all patches across frames to capture inter-frame contextual dynamics in a temporally causal manner.
\end{itemize}

Alternating these two paths across layers yields explicit spatio-temporal decoupling: the bidirectional path stabilizes static semantics, while the unidirectional path integrates contextual dynamics. This matches \textit{FPRL}’s hierarchical semantic modeling mechanism and provides a semantic information basis for subsequent lesion-centric static representations.

% -------------------------------------------------------------
% \paragraph{Teacher-Prior Adaptive Masking.}
\noindent\textbf{Teacher-Prior Adaptive Masking.} 
% To sharpen static semantic focus on the current view, we propose a Teacher-Prior Adaptive Masking (TPAM) as shown in Fig.~\ref{fig:FPRL}(b) that fuses a teacher-derived saliency prior with a lightweight attention head on tokens, thereby selecting visually prominent, lesion-correlated patches as visible while suppressing low-saliency background tokens.
To sharpen static semantic focus on the current view, we propose Teacher-Prior Adaptive Masking (TPAM), as shown in Fig.~\ref{fig:FPRL}(b), which fuses a teacher-derived saliency prior with a lightweight token attention head, thereby retaining visually salient and informative patches (e.g., lesion cues) as visible while suppressing low-saliency background tokens.

We compute a saliency prior $H \in \mathbb{R}^{N}$ through $\ell_2$-normalization of the teacher feature $z_t$, enhancing the selection of lesion-related tokens and suppressing background clutter. In parallel, a lightweight multi-head self-attention module applied to $\mathcal{V}_{\text{c}}$ with a linear projection produces logits that capture complementary, image-specific saliency to calibrate the final mask.
\begin{equation}
R = \mathrm{Linear}\!\big(\mathrm{MHA}(\mathrm{Embed}(\mathcal{V}_{\text{c}}))\big) \in \mathbb{R}^{N},
\end{equation}

We fuse the two cues and compute patch-wise sampling probabilities:
\begin{equation}
S = \alpha H + (1-\alpha) R, \qquad
P = \mathrm{Softmax}(S),
\end{equation}
where $\alpha \in [0,1]$ balances the teacher prior and the attention-derived mask logits. 
Visible patches are then selected by Top-$K$ over $P$, yielding a binary (learnable) mask $M \in \{0,1\}^{N}$ with visible index set $\Omega_v$ and masked index set $\Omega_m$.
The masked current view is fed into the student encoder to obtain the current representation:
\begin{equation}
z_c = f_s\!\big(X_{\!c}^{(\Omega_v)}\big) \in \mathbb{R}^{|\Omega_v| \times d},
\end{equation}
where $X_{\!c}^{(\Omega_v)}$ denotes the subset of tokens selected by $M$. 
Since $P$ is parameterized by the attention head (and $\alpha$), the mask $M$ is learnable via the probabilities \cite{bandara2023adamae}, enabling TPAM to concentrate representation capacity on lesion-related local semantics while suppressing redundant background responses.

\begin{table*}[t]
\caption{Comparison with other latest SOTA methods on 3 downstream tasks. We report F1 score (\%) for PolypDiag, Dice (\%) for CVC-12k, and F1 score (\%) for KUMC, respectively.}
\small
\centering
\setlength{\tabcolsep}{10pt}
\renewcommand{\arraystretch}{1.2}
\begin{tabular}{ccccccc}
\toprule
Method              & Venue   & Year & \begin{tabular}[c]{@{}c@{}}Pretrain\\  Time(h)\end{tabular} & \begin{tabular}[c]{@{}c@{}}PolypDiag\\ (Classification)\end{tabular} & \begin{tabular}[c]{@{}c@{}}CVC-12k\\ (Segmentation)\end{tabular} & \begin{tabular}[c]{@{}c@{}}KUMC\\ (Detection)\end{tabular} \\ \hline
Scratch (Rand.init.) & -       & -    & N/A                                                             & 83.5 ± 1.3                                                            & 53.2 ± 3.2                                                         & 73.5 ± 4.3                                                   \\ \hline
FAME \cite{FAME}               & CVPR    & 2022 & 48.9                                                            & 85.4 ± 0.8                                                            & 67.2 ± 1.3                                                         & 76.9 ± 1.2                                                   \\
ProViCo \cite{ProViCo}            & CVPR    & 2022 & 71.2                                                            & 86.9 ± 0.5                                                            & 69.0 ± 1.5                                                         & 78.6 ± 1.7                                                   \\
VCL \cite{VCL}                & ECCV    & 2022 & 74.9                                                            & 87.6 ± 0.6                                                            & 69.1 ± 1.2                                                         & 78.1 ± 1.9                                                   \\
ST-Adapter \cite{ST-Adapter}         & NeurIPS & 2022 & 8.1                                                             & 84.8 ± 0.7                                                            & 64.3 ± 1.9                                                         & 74.9 ± 2.9                                                   \\
VideoMAE \cite{tong2022videomae}           & NeurIPS & 2022 & 25.3                                                            & 91.4 ± 0.8                                                            & 80.9 ± 1.0                                                         & 82.8 ± 1.9                                                   \\
Endo-FM \cite{wang2023foundationEndoFM}            & MICCAI  & 2023 & 20.4                                                            & 90.7 ± 0.4                                                            & 73.9 ± 1.2                                                         & 84.1 ± 1.3                                                   \\
DropMAE \cite{wu2023dropmae}            & CVPR    & 2023 & 37.9                                                            & 88.2 ± 0.8                                                            & 80.9 ± 0.3                                                         & 81.7 ± 2.6                                                   \\
VideoMAE V2 \cite{wang2023videomaev2}        & CVPR    & 2023 & 17.3                                                            & 89.6 ± 1.4                                                            & 81.0 ± 0.4                                                         & 84.2 ± 1.0                                                   \\
M$^2$CRL \cite{hu2024mmcrl}             & NeurIPS & 2024 & 24.3                                                               & 94.2 ± 0.7                                                                    & 81.4 ± 0.8                                                                 & 86.3 ± 0.8                                                             \\
VideoMamba \cite{li2024videomamba}       & ECCV    & 2024 & 55.4                                                               & 88.2 ± 0.9                                                                    & 82.8 ± 0.5                                                                  & 84.1 ± 1.0                                                             \\
EndoFM-LV \cite{wang2025improvingEndoFM-LV}         & JBHI    & 2025 & 56.1                                                               & 94.5 ± 0.4                                                                    & 82.6 ± 0.3                                                                 & 72.6 ± 1.2                                                            \\
EndoMamba \cite{tian2025endomamba}      & MICCAI  & 2025 & 38.2                                                               & 94.5 ± 0.2                                                                   & 84.5 ± 0.8                                                                 & 88.8 ± 0.1                                                            \\ 
\hline
FPRL                 & Ours    & -    & 18.2                                                            & \textbf{95.2 ± 0.3}{$\color{highlight}\textbf{\footnotesize{$\uparrow$}0.7}$}                                                   & \textbf{86.1 ± 0.1}{$\color{highlight}\textbf{\footnotesize{$\uparrow$}1.6}$}                                                & \textbf{89.8 ± 0.1}{$\color{highlight}\textbf{\footnotesize{$\uparrow$}1.0}$}                                             \\ 
\bottomrule
\end{tabular}
\label{tab:result}
\end{table*}

\subsection{Contextual Perception for Semantic Correspondence}
\label{3.3}

Building upon the static representation, we further propose \textit{Contextual Semantic Perception} to explicitly establish reliable correspondence across views and model structured inter-frame evolution. This is achieved by two dedicated submodules: Cross-View Masked Feature Completion (CVMFC) and Attention-Guided Temporal Prediction (AGTP).

% -------------------------------------------------------------
% \paragraph{Cross-View Masked Feature Completion.}
\noindent\textbf{Cross-View Masked Feature Completion.} 
As shown in Fig.~\ref{fig:FPRL}(d), CVMFC performs fine-grained correspondence in the latent space and completes the masked current stream by retrieving semantics from adjacent views. 
Given the masked current features $z_c\!\in\!\mathbb{R}^{N\times d}$ and the student features of the past and future views $z_p,z_f\!\in\!\mathbb{R}^{N\times d}$, we adopt a Transformer-style block (cross-attention $\rightarrow$ self-attention $\rightarrow$ FFN) with two symmetric paths that use $z_p$ and $z_f$ as a temporal dictionary. 
Queries are formed from the current view, while keys/values come from the adjacent views.
For example, we set queries from the current view and keys/values from the past view:
\begin{equation}
\label{eq:cvmfc_proj_p}
Q_c = z_c W_q,\qquad
K_p = z_p W_k,\qquad
V_p = z_p W_v ,
\end{equation}
and compute cross-attention:
\begin{equation}
\label{eq:cvmfc_cross_p}
z_c^{\prime}=\mathrm{Softmax}\!\left(\frac{Q_c K_p^\top}{\sqrt{d}}\right)V_p ,
\end{equation}
The result is refined by self-attention and FFN:
\begin{equation}
\label{eq:cvmfc_final_p}
z_c^{\prime\prime}=\mathrm{SA}(z_c^{\prime}),\qquad 
z_c^{p}=\mathrm{FFN}(z_c^{\prime\prime}) ,
\end{equation}
Similarly, the $z_c^f$ can be generated in the same manner.

Layer normalization and residual connections follow standard practice (omitted for brevity). The outputs $z_c^p$ and $z_c^f$ are separately aligned with the frozen teacher features $z_t$ of the current view in Section~\ref{3.4}. 
This module establishes cross-view correspondence in a compact latent space and preserves temporal semantic continuity at the token level.

% -------------------------------------------------------------
% \paragraph{Attention-Guided Temporal Prediction.}
\noindent\textbf{Attention-Guided Temporal Prediction.}
As shown in Fig.~\ref{fig:FPRL}(c), AGTP enforces view-level temporal correspondence consistency across different views. 
Unlike CVMFC’s token-wise retrieval, AGTP aggregates each view according to its cross-attention with the current view and compares it to a lesion-centric current representation produced by \emph{Static Semantic Focus} (Section~\ref{3.2}). 

Let $A_c^{p}\!\in\!\mathbb{R}^{N\times N}$ be the cross-attention map from the current view (queries) to the past view (keys/values) computed in CVMFC, and let $z_p=\{z_p^{(j)}\}_{j=1}^{N}$ be the past-view tokens. 
We first derive normalized pooling weights from $A_c^{p}$, then pool $z_p$, and project the pooled vector by a target head:
\begin{equation}
\label{eq:agtp_past}
\left\{
\begin{aligned}
a_p  &\,=\, \mathrm{Norm}\!\left(\tfrac{1}{N}\sum_{i=1}^{N} A_c^{p}(i,:)\right) \in \mathbb{R}^{N},\\
z_p' &\,=\, \sum_{j=1}^{N} a_p(j)\, z_p^{(j)} \in \mathbb{R}^{d},\\
p_p  &\,=\, \phi_t\!\big(z_p'\big) \in \mathbb{R}^{d'},
\end{aligned}
\right.
\end{equation}
where $\mathrm{Norm}(u)=u/(\mathbf{1}^\top u)$ ensures $\sum_j a_p(j)=1$, and $\phi_t$ is a lightweight projector (\textit{e.g.}, two linear layers with nonlinearity). 
The future view is handled symmetrically using $A_c^{f}$ and $z_f$ to obtain $p_f$.

% For the current view, the visual patches revolve around the lesion-related region, so we omit attention pooling and directly project a global average over current tokens:
For the current view, the visual patches preserve local evidence (e.g., lesion cues), so we directly apply global average pooling to the current tokens:

\begin{equation}
\label{eq:agtp_cur}
\left\{
\begin{aligned}
&\mathrm{GAP}(z_c) \,=\, \tfrac{1}{N}\sum_{i=1}^{N} z_c^{(i)} \in \mathbb{R}^{d},\\
&p_c \,=\, \phi_c\!\big(\mathrm{GAP}(z_c)\big) \in \mathbb{R}^{d'},
\end{aligned}
\right.
\end{equation}

To provide stable, non-degenerate targets, the target head is updated from the query head by EMA \cite{grill2020BYOL} with momentum $m\!\in\![0,1)$ (stop-gradient on $p_p$ and $p_f$):
\begin{equation}
\label{eq:agtp_ema}
\phi_t \;\leftarrow\; m\,\phi_t \;+\; (1{-}m)\,\phi_c,
\end{equation}

The view-level objectives using $(p_c,p_p,p_f)$ are defined in Section~\ref{3.4}.

\subsection{Pre-training Objective}
\label{3.4}

We jointly optimize three objectives to couple static semantics and contextual semantics: 
1) pixel-level reconstruction on masked current tokens to recover lesion-centric texture/boundary details and suppress low-level motion noise, 
2) cross-view feature alignment between CVMFC outputs and the frozen teacher to establish fine-grained temporal correspondence, and 
3) temporal prediction contrastive learning from AGTP to maintain temporal consistency of view-level semantics.

% \paragraph{Pixel Reconstruction.}
\noindent\textbf{Pixel Reconstruction.} Normalized pixel values are used as explicit signals to stabilize the pre-training process.
Given the tokenized current view $X_c$, with the masked index set $\Omega_m$ from TPAM, and the reconstruction $\hat{X}_c$ from a lightweight decoder $g$ using $z_c$ as input. 
We minimize the mean squared error over masked tokens as:
\begin{equation}
\label{eq:loss_rec}
\mathcal{L}_{\mathrm{Rec}}
 \;=\; \frac{1}{|\Omega_m|}\sum_{i\in\Omega_m}
 \big\|\,\hat{X}_c^{(i)} - X_c^{(i)}\,\big\|_2^2,
\end{equation}

% \paragraph{Cross-View Feature Alignment.}
\noindent\textbf{Cross-View Feature Alignment.} CVMFC yields two updated current streams $z_c^p$ and $z_c^f$, which are aligned to the frozen teacher features $z_t$ on the current view. 
We use cosine similarity for past/future-to-teacher alignment and an $\ell_2$ consistency between the two streams for stable features completion:
\begin{equation}
\label{eq:loss_pt}
\mathcal{L}_{pt}
= 1 - \frac{1}{|\Omega_m|} \sum_{i\in\Omega_m}
\cos\!\big(z_c^{p(i)},\, z_t^{(i)}\big),
\end{equation}

\begin{equation}
\label{eq:loss_ft}
\mathcal{L}_{ft}
= 1 - \frac{1}{|\Omega_m|} \sum_{i\in\Omega_m}
\cos\!\big(z_c^{f(i)},\, z_t^{(i)}\big),
\end{equation}

\begin{equation}
\label{eq:loss_pf}
\mathcal{L}_{pf}
= \frac{1}{|\Omega_m|} \sum_{i\in\Omega_m}
\big\| z_c^{p(i)} - z_c^{f(i)} \big\|_2^2,
\end{equation}
where 
\(\cos(u,v)=\tfrac{u^\top v}{\|u\|_2\|v\|_2}\).
The alignment loss is as follows:
\begin{equation}
\label{eq:loss_align}
\mathcal{L}_{\mathrm{Align}} \;=\; 
\mathcal{L}_{pt} \;+\; \mathcal{L}_{ft} \;+\; \lambda_{\mathrm{pf}}\, \mathcal{L}_{pf},
\end{equation}
where $\lambda_{\mathrm{pf}}$ is the scalable weight.

% \paragraph{Temporal Prediction Contrastive Learning.}
\noindent\textbf{Temporal Prediction Contrastive Learning.} AGTP produces the view-level embeddings \(p_c\), \(p_p\) and  \(p_f\). 
We adopt InfoNCE with in-batch keys, positives are \((p_c, p_p)\) and \((p_c, p_f)\), while negatives are other keys in the batch, thus the contrastive learning loss can be defined with temperature $\tau$ and key set $\mathcal{K}$ as:
\begin{equation}
\label{eq:loss_cl}
\begin{aligned}
\mathcal{L}_{\mathrm{CL}}
= &- \log \frac{\exp(\langle p_c, p_p\rangle/\tau)}
{\sum_{k\in\mathcal{K}} \mathbf{1}[k\neq p_p] \exp(\langle p_c, k\rangle/\tau)} \\
&- \log \frac{\exp(\langle p_c, p_f\rangle/\tau)}
{\sum_{k\in\mathcal{K}} \mathbf{1}[k\neq p_f] \exp(\langle p_c, k\rangle/\tau)},
\end{aligned}
\end{equation}

% \paragraph{Overall Objective.}
\noindent\textbf{Overall Objective.} The total pre-training loss is a weighted sum, which can be formulated as:
\begin{equation}
\label{eq:loss_total}
\mathcal{L}_{\mathrm{total}} 
\;=\; \lambda_{\mathrm{1}}\,\mathcal{L}_{\mathrm{Rec}}
\;+\; \lambda_{\mathrm{2}}\,\mathcal{L}_{\mathrm{Align}}
\;+\; \lambda_{\mathrm{3}}\,\mathcal{L}_{\mathrm{CL}}.
\end{equation}
where $\lambda_{\mathrm{1}}, \lambda_{\mathrm{2}}$ and $\lambda_{\mathrm{3}}$ are scalar weights. In combination, these objectives bind static, lesion-centric semantics and temporal context into a unified hierarchical representation.

\begin{figure*}
    \centering
    \includegraphics[width=1\linewidth]{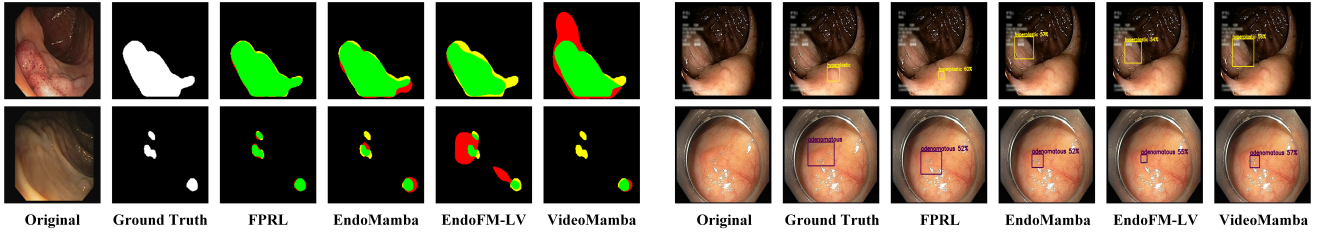}
    \caption{Qualitative results of segmentation and detection tasks. The segmentation results (green, red, and yellow regions represent the true positive, false positive, and false negative, respectively) on the left are from the CVC-12k dataset, while the detection results on the right are from the KUMC dataset.}
    \label{fig:qualitative}
\end{figure*}

\begin{table}[t]
\caption{Ablation on different components.}
\centering
\footnotesize
\renewcommand{\arraystretch}{1.25}
\setlength{\tabcolsep}{3.8pt}
\begin{adjustbox}{width=\linewidth}
\begin{tabular}{cc|ccc|ccc}
\hline
\multirow{2}{*}{$\mathcal{L}_{Rec}$} & 
\multirow{2}{*}{$\mathcal{L}_{CL}$} & 
\multicolumn{3}{c|}{$\mathcal{L}_{Align}$} & 
\multicolumn{3}{c}{Performance (\%)} \\ \cline{3-5} \cline{6-8}
 &  & $\mathcal{L}_{pt}$ & $\mathcal{L}_{ft}$ & $\mathcal{L}_{pf}$ & Cla. & Seg. & Det. \\ \hline
 &  & \checkmark &  &  & 92.3 $\pm$ 0.4 & 83.8 $\pm$ 1.1 & 84.0 $\pm$ 1.3 \\
 &  &  & \checkmark &  & 92.2 $\pm$ 0.5 & 83.5 $\pm$ 1.2 & 83.5 $\pm$ 1.7 \\
 &  & \checkmark & \checkmark &  & 94.1 $\pm$ 0.8 & 85.7 $\pm$ 1.1 & 87.3 $\pm$ 0.8 \\ 
 &  & \checkmark & \checkmark & \checkmark & 94.2 $\pm$ 0.5 & 85.5 $\pm$ 0.2 & 88.5 $\pm$ 0.3 \\
\checkmark &  &  &  &  & 92.7 $\pm$ 0.8 & 84.3 $\pm$ 0.2 & 85.4 $\pm$ 0.4 \\
 & \checkmark &  &  &  & 93.0 $\pm$ 1.4 & 79.6 $\pm$ 1.2 & 84.1 $\pm$ 1.3 \\
\checkmark &  & \checkmark & \checkmark & \checkmark  & 93.8 $\pm$ 0.5 & 85.8 $\pm$ 0.3 & 87.5 $\pm$ 0.3 \\
 & \checkmark & \checkmark & \checkmark & \checkmark  & 94.2 $\pm$ 0.4 & 84.0 $\pm$ 1.5 & 86.1 $\pm$ 0.9 \\
\rowcolor{pink!28}
\checkmark & \checkmark & \checkmark & \checkmark & \checkmark  & \textbf{95.2 $\pm$ 0.3}  & \textbf{86.1 $\pm$ 0.1}  & \textbf{89.8 $\pm$ 0.1} \\ \hline
\end{tabular}
\end{adjustbox}
\label{tab:Components}
\end{table}

\section{Experiments}
\label{sec:experiment}

%-------------------------------------------------------------------------
\subsection{Datasets and Experimental Settings}
\label{4.1}

\noindent{\textbf{Datasets.}} We conduct experiments on 11 publicly available endoscopic video datasets: Colonoscopic \cite{mesejo2016Colonoscopic}, SUN-SEG \cite{ji2022videopolyp}, LDPolypVideo \cite{ma2021ldpolypvideo}, Hyper-Kvasir \cite{borgli2020hyperkvasir}, Kvasir-Capsule \cite{smedsrud2021kvasir}, CholecTriplet \cite{nwoye2022CholecTriplet}, Renji-Hospital \cite{wang2023foundationEndoFM}, PolypDiag \cite{tian2022PolypDiag}, CVC-12k \cite{bernal2015CVC-12k}, KUMC \cite{li2021KUMC} and Cholec80 \cite{twinanda2016cholec80}. These comprise 33,311 videos totaling approximately 6M frames, covering 3 types of endoscopy examination protocols and 10$+$ different diseases.
% The videos are processed into 30fps clips with an average duration of 5 seconds.
For pre-training, the raw videos are temporally segmented into 30 fps clips with an average duration of 5 seconds.

\noindent{\textbf{Settings.}} The first 7 datasets are used for pre-training and we sample 3 views from each video where frames are resized to spatial size of 224×224 and frame number of 2 as input. We adopt EndoMamba-S \cite{tian2025endomamba} with a patch size of 16 as the backbone model in our framework. The model is trained using the AdamW optimizer with a base learning rate of 1.5e-4, a cosine learning rate schedule for
400 epochs, and a batch size of 64, with the first five epochs dedicated to linear warmup. For feature alignment, the pretrained VideoMamba-S \cite{li2024videomamba} serves as the teacher model, with the loss weight set to \(\lambda_{\mathrm{2}}\) = 0.8 and other hyper-parameters of the loss function are set as follows: \(\lambda_{\mathrm{1}}\) = 1.0, \(\lambda_{\mathrm{3}}\) = 1.0, and \(\lambda_{\mathrm{pf}}\) = 20 based on preliminary experiments.

\noindent{\textbf{Evaluations.}} We evaluate our framework on 4 downstream tasks: classification on PolypDiag, segmentation on CVC-12k, detection on KUMC, and recognition on Cholec80. 
Surgical phase recognition (Cholec80) is provided in \textit{Section C.1 of Supplementary Material}. This experiment follows a standardized protocol and emphasizes the generalizability of our method to long-horizon recognition.

%-------------------------------------------------------------------------
\subsection{Comparison with Prior Works}
\label{4.2}

We compare our method with several recent state-of-the-art (SOTA) approaches for endoscopy video analysis. The results of these methods are taken from the comparative method reported in M$^2$CRL \cite{hu2024mmcrl}, including FAME \cite{FAME}, ProViCo \cite{ProViCo}, VCL \cite{VCL}, ST-Adapter \cite{ST-Adapter}, VideoMAE \cite{tong2022videomae}, Endo-FM \cite{wang2023foundationEndoFM}, DropMAE \cite{wu2023dropmae},
and VideoMAE V2 \cite{wang2023videomaev2}. Additionally, we also compare the latest methods for video self-supervised and endoscopic video representation learning: VideoMamba \cite{li2024videomamba}, EndoFM-LV \cite{wang2025improvingEndoFM-LV}, and EndoMamba \cite{tian2025endomamba}. For a fair comparison, all methods are pretrained on the same union of 7 datasets as our \textit{FPRL}. All experimental settings are referred to those documented in their original papers or in their released codes.

\noindent{\textbf{Quantitative Evaluation.}} From Table \ref{tab:result}, we can observe that our method achieves superior performance compared to SOTA methods. Notably, under the same model architecture, our \textit{FPRL} outperforms the best existing method, EndoMamba, by improving 0.7\% F1 in classification, 1.6\% Dice in segmentation, and 1.0\% F1 in detection tasks, while also reducing pre-training time by 52\%. 
These advancements can be attributed to our method, which improves representation learning by introducing a hierarchical semantic modeling mechanism that utilizes multi-view sparse sampling.

\noindent{\textbf{Qualitative Evaluation.}} 
As visualized in Fig.~\ref{fig:qualitative}, our method yields visually convincing results in both segmentation and detection tasks. For segmentation, \textit{FPRL} effectively delineates polyps of varying sizes, thanks to the teacher-prior adaptive masking strategy that enhances focus on lesion-related regions. Furthermore, in detection tasks, it obtains robust performance, adeptly managing challenging scenarios characterized by complex backgrounds.

%-------------------------------------------------------------------------
\subsection{Ablation Study}
\label{4.3}

\begin{table}[t]
\caption{Ablation on different masking strategies.}
\centering
\footnotesize
\renewcommand{\arraystretch}{1.25}
\setlength{\tabcolsep}{5pt}
\begin{adjustbox}{width=\linewidth}
\begin{tabular}{ccc|ccc}
\hline
\multicolumn{3}{c|}{Masking Strategy} & \multicolumn{3}{c}{Performance (\%)} \\ \cline{1-3} \cline{4-6}
Random & Adaptive & Teacher-Prior & Cla. & Seg. & Det. \\ \hline
\checkmark &  &  & 93.8 $\pm$ 0.5 & 85.6 $\pm$ 0.9 & 87.8 $\pm$ 1.1 \\
 & \checkmark &  & 94.5 $\pm$ 0.4 & 85.6 $\pm$ 0.7 & 83.9 $\pm$ 1.0 \\
 &  & \checkmark & 93.8 $\pm$ 0.3 & 85.7 $\pm$ 0.2 & 86.7 $\pm$ 0.8 \\
\rowcolor{pink!28}
 & \checkmark & \checkmark & \textbf{95.2 $\pm$ 0.3}  & \textbf{86.1 $\pm$ 0.1}  & \textbf{89.8 $\pm$ 0.1} \\ \hline
\end{tabular}
\end{adjustbox}
\label{tab:mask_ablation}
\end{table}
% \vspace{-0.5em}  % 缩小表格后的间距

\noindent{\textbf{Analysis of Components.}}
Results in rows 1-3 of Table \ref{tab:Components} reveal that latent-space feature alignment enables the student to inherit the teacher's representations. When masked completion (\(\mathcal{L}_{pt}+\mathcal{L}_{ft}\)) is applied to both paths, performance rises by roughly 1.9\%/2.0\%/3.6\% over single-path alignment. Subsequently, enforcing past–future consistency (\(\mathcal{L}_{pf}\)) further boosts 1.2\% in detection and reduces output fluctuation, leading to a more consistent reconstruction space.
The consistent gains from pixel reconstruction (\(\mathcal{L}_{\mathrm{Rec}}\)) are further maximized by the temporal contrastive loss \((\mathcal{L}_{\mathrm{CL}}\)), which provides an additional performance increase of 1.4\%/0.3\%/2.3\%. This demonstrates that their combination provides the stable features needed to enhance multi-view collaboration, with the full objective exceeding the best alignment-only version by 1.0\%/0.6\%/1.3\%.

\noindent{\textbf{Masking Strategy.}}
Table \ref{tab:mask_ablation} compares different masking strategies under a fixed masking ratio. Using random masking already surpasses other leading methods in segmentation, establishing a strong baseline. While adaptive masking improves classification, its detection performance remains limited, suggesting difficulty in identifying truly relevant regions. Teacher-prior guidance aligns with this strategy, yet the best results come from combining both approaches, demonstrating that integrating both information sources enables the model to better capture lesion-related features.

\begin{table}[t]
\caption{Ablation on different mask ratios.}
\centering
\footnotesize
\renewcommand{\arraystretch}{1.0}
\setlength{\tabcolsep}{6pt}
\begin{adjustbox}{width=\linewidth}
\begin{tabular}{c|ccc}
\hline
\multirow{2}{*}{Mask Ratio (\%)} & \multicolumn{3}{c}{Performance (\%)} \\
\cline{2-4}
& Cla. & Seg. & Det. \\ \hline
70 & 93.0 $\pm$ 2.7 & 85.3 $\pm$ 0.2 & 87.3 $\pm$ 1.2 \\
75 & 94.4 $\pm$ 0.5 & 85.1 $\pm$ 0.8 & 83.7 $\pm$ 1.0 \\
80 & 95.1 $\pm$ 0.3 & 85.4 $\pm$ 0.2 & 86.7 $\pm$ 0.9 \\
85 & 94.5 $\pm$ 0.6 & 85.5 $\pm$ 0.2 & 86.7 $\pm$ 0.7 \\
\rowcolor{pink!28}
90 & \textbf{95.2 $\pm$ 0.3}  & \textbf{86.1 $\pm$ 0.1}  & \textbf{89.8 $\pm$ 0.1} \\
95 & 92.7 $\pm$ 0.4 & 78.5 $\pm$ 1.4 & 87.1 $\pm$ 1.5 \\ \hline
\end{tabular}
\end{adjustbox}
\label{tab:mask_ratio}
\end{table}

\begin{table}[t]
\caption{Ablation on AGTP module configurations.}
\centering
\footnotesize
\renewcommand{\arraystretch}{1.2}
\setlength{\tabcolsep}{5.2pt}
\begin{adjustbox}{width=\linewidth}
\begin{tabular}{cc|ccc}
\hline
\multicolumn{2}{c|}{AGTP Configuration} & \multicolumn{3}{c}{Performance (\%)} \\ \cline{1-2} \cline{3-5}
Attn Pool & EMA Update & Cla. & Seg. & Det. \\ \hline
 &  & 93.0 $\pm$ 0.6 & 86.0 $\pm$ 0.3 & 86.9 $\pm$ 0.2 \\   
\checkmark &  & 95.0 $\pm$ 0.5 & 86.0 $\pm$ 0.1 & 86.4 $\pm$ 0.1 \\   
 & \checkmark & 95.2 $\pm$ 0.4 & 86.0 $\pm$ 0.3 & 87.7 $\pm$ 0.1 \\
\rowcolor{pink!28}   
\checkmark & \checkmark & \textbf{95.2 $\pm$ 0.3}  & \textbf{86.1 $\pm$ 0.1}  & \textbf{89.8 $\pm$ 0.1} \\  
\hline
\end{tabular}
\end{adjustbox}
\label{tab:agtp_ablation}
\end{table}

\noindent{\textbf{Masking Ratio.}}
Table~\ref{tab:mask_ratio} summarizes the impact of varying masking ratios. We can observe that as the ratio increases from 70\% to 90\%, performance across the three downstream tasks fluctuates slightly but generally improves. The increased masking encourages the model to learn more effectively from fewer visible patches, thereby enhancing its ability to capture temporal correspondences. However, at an extreme ratio of 95\%, the limited visible context hinders learning, leading to suboptimal results.

\noindent{\textbf{AGTP Configurations.}}
Our AGTP module implements temporal contrastive learning by deriving view-level features from pooled patch tokens. Within this module, attention pooling highlights salient regions, while EMA updating stabilizes the learning of representations. As presented in Table~\ref{tab:agtp_ablation}, combining both mechanisms yields the most significant gains for contrastive learning. Please refer to \textit{Section C.2 of Supplementary Material} for more ablation analyses.

\section{Conclusion}
\label{sec:conclusion}

% In this study, we propose a cognitive framework called \textit{FPRL} for endoscopic video analysis, which employs hierarchical semantic modeling to decouple and jointly learn both static and contextual semantics. Concretely, we propose a teacher-prior adaptive masking strategy to emphasize lesion-centric regions and reduce redundant temporal dependencies. This is coupled with cross-view masked feature completion and attention-guided temporal prediction to capture structural and temporal patterns in semantic regions. Such design effectively mitigates motion bias and produces robust representations for endoscopic videos. Extensive experiments conducted on multiple publicly available datasets verify that our \textit{FPRL} consistently outperforms existing methods across diverse downstream tasks.
In this study, we propose \textit{FPRL}, a cognitive framework for endoscopic video analysis that hierarchically decouples and jointly learns static and contextual semantics. Specifically, a teacher-prior adaptive masking strategy emphasizes semantic regions while mitigating temporal redundancy, alongside cross-view masked feature completion and attention-guided temporal prediction to capture structural and temporal patterns. These designs effectively mitigate motion bias and yield robust representations. Extensive experiments on multiple public datasets show that \textit{FPRL} consistently outperforms existing methods across diverse downstream tasks.

%For Camera Ready
\noindent{\textbf{Limitation and Future Work.}}
We explored a single-frame pre-training variant to better leverage the semantic advantages of sparse views, but its performance is constrained by noisy supervision from common endoscopic artifacts. Future work will investigate quality-aware sampling and lightweight multi-frame designs to improve robustness.

\section*{Acknowledgments}
This work was supported in part by the National Natural Science Foundation of China under Grants 62502419, 62272404, 62372170, 62376238, and 12571591, and in part by the Research Foundation of Education Department of Hunan Province of China under Grant 23A0146.

% \clearpage
{
    \small
    \bibliographystyle{ieeenat_fullname}
    \bibliography{main}

@inproceedings{hirsch2023selfendoscopic,
  title={Self-supervised learning for endoscopic video analysis},
  author={Hirsch, Roy and Caron, Mathilde and Cohen, Regev and Livne, Amir and Shapiro, Ron and Golany, Tomer and Goldenberg, Roman and Freedman, Daniel and Rivlin, Ehud},
  booktitle={International Conference on Medical Image Computing and Computer-Assisted Intervention},
  pages={569--578},
  year={2023},
  organization={Springer}
}

@article{ali2021deependoscopy,
  title={A deep learning framework for quality assessment and restoration in video endoscopy},
  author={Ali, Sharib and Zhou, Felix and Bailey, Adam and Braden, Barbara and East, James E and Lu, Xin and Rittscher, Jens},
  journal={Medical Image Analysis},
  volume={68},
  pages={101900},
  year={2021},
  publisher={Elsevier}
}

@article{ji2022videopolyp,
  title={Video polyp segmentation: A deep learning perspective},
  author={Ji, Ge-Peng and Xiao, Guobao and Chou, Yu-Cheng and Fan, Deng-Ping and Zhao, Kai and Chen, Geng and Van Gool, Luc},
  journal={Machine Intelligence Research},
  volume={19},
  number={6},
  pages={531--549},
  year={2022},
  publisher={Springer}
}

@article{tong2022videomae,
  title={Videomae: Masked autoencoders are data-efficient learners for self-supervised video pre-training},
  author={Tong, Zhan and Song, Yibing and Wang, Jue and Wang, Limin},
  journal={Advances in Neural Information Processing Systems},
  volume={35},
  pages={10078--10093},
  year={2022}
}

@inproceedings{bertasius2021spaceTimesformer,
  title={Is space-time attention all you need for video understanding?},
  author={Bertasius, Gedas and Wang, Heng and Torresani, Lorenzo},
  booktitle={International Conference on Machine Learning},
  volume={2},
  number={3},
  pages={4},
  year={2021}
}

@inproceedings{wang2023foundationEndoFM,
  title={Foundation model for endoscopy video analysis via large-scale self-supervised pre-train},
  author={Wang, Zhao and Liu, Chang and Zhang, Shaoting and Dou, Qi},
  booktitle={International Conference on Medical Image Computing and Computer-Assisted Intervention},
  pages={101--111},
  year={2023},
  organization={Springer}
}

@inproceedings{wang2023videomaev2,
  title={Videomae v2: Scaling video masked autoencoders with dual masking},
  author={Wang, Limin and Huang, Bingkun and Zhao, Zhiyu and Tong, Zhan and He, Yinan and Wang, Yi and Wang, Yali and Qiao, Yu},
  booktitle={Proceedings of the IEEE/CVF Conference on Computer Vision and Pattern Recognition},
  pages={14549--14560},
  year={2023}
}

@article{gupta2023siameseMAE,
  title={Siamese masked autoencoders},
  author={Gupta, Agrim and Wu, Jiajun and Deng, Jia and Li, Fei-Fei},
  journal={Advances in Neural Information Processing Systems},
  volume={36},
  pages={40676--40693},
  year={2023}
}

@inproceedings{pei2024videomac,
  title={Videomac: Video masked autoencoders meet convnets},
  author={Pei, Gensheng and Chen, Tao and Jiang, Xiruo and Liu, Huafeng and Sun, Zeren and Yao, Yazhou},
  booktitle={Proceedings of the IEEE/CVF Conference on Computer Vision and Pattern Recognition},
  pages={22733--22743},
  year={2024}
}

@inproceedings{liu2025futureTCoRe,
  title={When the future becomes the past: Taming temporal correspondence for self-supervised video representation learning},
  author={Liu, Yang and Xu, Qianqian and Wen, Peisong and Dai, Siran and Huang, Qingming},
  booktitle={Proceedings of the Computer Vision and Pattern Recognition Conference},
  pages={24033--24044},
  year={2025}
}

@article{wang2025improvingEndoFM-LV,
  title={Improving Foundation Model for Endoscopy Video Analysis via Representation Learning on Long Sequences},
  author={Wang, Zhao and Liu, Chang and Zhu, Lingting and Wang, Tongtong and Zhang, Shaoting and Dou, Qi},
  journal={IEEE Journal of Biomedical and Health Informatics},
  year={2025},
  publisher={IEEE}
}

@article{hu2024mmcrl,
  title={Multi-view masked contrastive representation learning for endoscopic video analysis},
  author={Hu, Kai and Xiao, Ye and Zhang, Yuan and Gao, Xieping},
  journal={Advances in Neural Information Processing Systems},
  volume={37},
  pages={47987--48014},
  year={2024}
}

@inproceedings{tian2025endomamba,
  title={Endomamba: An efficient foundation model for endoscopic videos via hierarchical pre-training},
  author={Tian, Qingyao and Liao, Huai and Huang, Xinyan and Yang, Bingyu and Lei, Dongdong and Ourselin, Sebastien and Liu, Hongbin},
  booktitle={International Conference on Medical Image Computing and Computer-Assisted Intervention},
  pages={224--234},
  year={2025},
  organization={Springer}
}

@inproceedings{li2024videomamba,
  title={Videomamba: State space model for efficient video understanding},
  author={Li, Kunchang and Li, Xinhao and Wang, Yi and He, Yinan and Wang, Yali and Wang, Limin and Qiao, Yu},
  booktitle={European Conference on Computer Vision},
  pages={237--255},
  year={2024},
  organization={Springer}
}

@inproceedings{actionrec2021large-scale,
  title={A large-scale study on unsupervised spatiotemporal representation learning},
  author={Feichtenhofer, Christoph and Fan, Haoqi and Xiong, Bo and Girshick, Ross and He, Kaiming},
  booktitle={Proceedings of the IEEE/CVF Conference on Computer Vision and Pattern Recognition},
  pages={3299--3309},
  year={2021}
}

@inproceedings{jenni2021timeCorrespendence1,
  title={Time-equivariant contrastive video representation learning},
  author={Jenni, Simon and Jin, Hailin},
  booktitle={Proceedings of the IEEE/CVF International Conference on Computer Vision},
  pages={9970--9980},
  year={2021}
}

@inproceedings{qian2021timeCorrespendence2,
  title={Spatiotemporal contrastive video representation learning},
  author={Qian, Rui and Meng, Tianjian and Gong, Boqing and Yang, Ming-Hsuan and Wang, Huisheng and Belongie, Serge and Cui, Yin},
  booktitle={Proceedings of the IEEE/CVF Conference on Computer Vision and Pattern Recognition},
  pages={6964--6974},
  year={2021}
}

@inproceedings{sermanet2018timeCorrespendence3,
  title={Time-contrastive networks: Self-supervised learning from video},
  author={Sermanet, Pierre and Lynch, Corey and Chebotar, Yevgen and Hsu, Jasmine and Jang, Eric and Schaal, Stefan and Levine, Sergey and Brain, Google},
  booktitle={2018 IEEE International Conference on Robotics and Automation (ICRA)},
  pages={1134--1141},
  year={2018},
  organization={IEEE}
}

@article{qing2023hierarchicalconsistency,
  title={Self-supervised learning from untrimmed videos via hierarchical consistency},
  author={Qing, Zhiwu and Zhang, Shiwei and Huang, Ziyuan and Xu, Yi and Wang, Xiang and Gao, Changxin and Jin, Rong and Sang, Nong},
  journal={IEEE Transactions on Pattern Analysis and Machine Intelligence},
  volume={45},
  number={10},
  pages={12408--12426},
  year={2023},
  publisher={IEEE}
}

@inproceedings{kuehne2011motionrec,
  title={HMDB: a large video database for human motion recognition},
  author={Kuehne, Hildegard and Jhuang, Hueihan and Garrote, Est{\'\i}baliz and Poggio, Tomaso and Serre, Thomas},
  booktitle={2011 International Conference on Computer Vision},
  pages={2556--2563},
  year={2011},
  organization={IEEE}
}

@article{kordopatis2019retrieval,
  title={FIVR: Fine-grained incident video retrieval},
  author={Kordopatis-Zilos, Giorgos and Papadopoulos, Symeon and Patras, Ioannis and Kompatsiaris, Ioannis},
  journal={IEEE Transactions on Multimedia},
  volume={21},
  number={10},
  pages={2638--2652},
  year={2019},
  publisher={IEEE}
}

@inproceedings{liu2024retrieval,
  title={Not all pairs are equal: Hierarchical learning for average-precision-oriented video retrieval},
  author={Liu, Yang and Xu, Qianqian and Wen, Peisong and Dai, Siran and Huang, Qingming},
  booktitle={Proceedings of the 32nd ACM International Conference on Multimedia},
  pages={3828--3837},
  year={2024}
}

@article{grill2020BYOL,
  title={Bootstrap your own latent-a new approach to self-supervised learning},
  author={Grill, Jean-Bastien and Strub, Florian and Altch{\'e}, Florent and Tallec, Corentin and Richemond, Pierre and Buchatskaya, Elena and Doersch, Carl and Avila Pires, Bernardo and Guo, Zhaohan and Gheshlaghi Azar, Mohammad and others},
  journal={Advances in Neural Information Processing Systems},
  volume={33},
  pages={21271--21284},
  year={2020}
}

@inproceedings{caron2021DINO,
  title={Emerging properties in self-supervised vision transformers},
  author={Caron, Mathilde and Touvron, Hugo and Misra, Ishan and J{\'e}gou, Herv{\'e} and Mairal, Julien and Bojanowski, Piotr and Joulin, Armand},
  booktitle={Proceedings of the IEEE/CVF International Conference on Computer Vision},
  pages={9650--9660},
  year={2021}
}

@article{oquab2023dinov2,
  title={Dinov2: Learning robust visual features without supervision},
  author={Oquab, Maxime and Darcet, Timoth{\'e}e and Moutakanni, Th{\'e}o and Vo, Huy and Szafraniec, Marc and Khalidov, Vasil and Fernandez, Pierre and Haziza, Daniel and Massa, Francisco and El-Nouby, Alaaeldin and others},
  journal={arXiv preprint arXiv:2304.07193},
  year={2023}
}

@inproceedings{chen2020SimCLR,
  title={A simple framework for contrastive learning of visual representations},
  author={Chen, Ting and Kornblith, Simon and Norouzi, Mohammad and Hinton, Geoffrey},
  booktitle={International Conference on Machine Learning},
  pages={1597--1607},
  year={2020},
  organization={PmLR}
}

@inproceedings{misra2016shuffleFrame,
  title={Shuffle and learn: unsupervised learning using temporal order verification},
  author={Misra, Ishan and Zitnick, C Lawrence and Hebert, Martial},
  booktitle={European Conference on Computer Vision},
  pages={527--544},
  year={2016},
  organization={Springer}
}

@inproceedings{lee2017shuffleFrame2,
  title={Unsupervised representation learning by sorting sequences},
  author={Lee, Hsin-Ying and Huang, Jia-Bin and Singh, Maneesh and Yang, Ming-Hsuan},
  booktitle={Proceedings of the IEEE International Conference on Computer Vision},
  pages={667--676},
  year={2017}
}

@inproceedings{benaim2020speednet,
  title={Speednet: Learning the speediness in videos},
  author={Benaim, Sagie and Ephrat, Ariel and Lang, Oran and Mosseri, Inbar and Freeman, William T and Rubinstein, Michael and Irani, Michal and Dekel, Tali},
  booktitle={Proceedings of the IEEE/CVF Conference on Computer Vision and Pattern Recognition},
  pages={9922--9931},
  year={2020}
}

@inproceedings{srivastava2015PredictFuture,
  title={Unsupervised learning of video representations using lstms},
  author={Srivastava, Nitish and Mansimov, Elman and Salakhudinov, Ruslan},
  booktitle={International Conference on Machine Learning},
  pages={843--852},
  year={2015},
  organization={PMLR}
}

@misc{
bardes2024vjepa,
title={V-{JEPA}: Latent Video Prediction for Visual Representation Learning},
author={Adrien Bardes and Quentin Garrido and Jean Ponce and Xinlei Chen and Michael Rabbat and Yann LeCun and Mido Assran and Nicolas Ballas},
year={2024},
howpublished = {\hypersetup{urlcolor=black}{\emph{\url{https://openreview.net/forum?id=WFYbBOEOtv}}}}
}

@inproceedings{he2022MAE,
  title={Masked autoencoders are scalable vision learners},
  author={He, Kaiming and Chen, Xinlei and Xie, Saining and Li, Yanghao and Doll{\'a}r, Piotr and Girshick, Ross},
  booktitle={Proceedings of the IEEE/CVF Conference on Computer Vision and Pattern Recognition},
  pages={16000--16009},
  year={2022}
}

@article{feichtenhofer2022ST-MAE,
  title={Masked autoencoders as spatiotemporal learners},
  author={Feichtenhofer, Christoph and Li, Yanghao and He, Kaiming and others},
  journal={Advances in Neural Information Processing Systems},
  volume={35},
  pages={35946--35958},
  year={2022}
}

@inproceedings{fan2023motionMGMAE,
  title={Motion-guided masking for spatiotemporal representation learning},
  author={Fan, David and Wang, Jue and Liao, Shuai and Zhu, Yi and Bhat, Vimal and Santos-Villalobos, Hector and MV, Rohith and Li, Xinyu},
  booktitle={Proceedings of the IEEE/CVF International Conference on Computer Vision},
  pages={5619--5629},
  year={2023}
}

@inproceedings{bandara2023adamae,
  title={Adamae: Adaptive masking for efficient spatiotemporal learning with masked autoencoders},
  author={Bandara, Wele Gedara Chaminda and Patel, Naman and Gholami, Ali and Nikkhah, Mehdi and Agrawal, Motilal and Patel, Vishal M},
  booktitle={Proceedings of the IEEE/CVF Conference on Computer Vision and Pattern Recognition},
  pages={14507--14517},
  year={2023}
}

@inproceedings{hernandez2024vicmae,
  title={Vic-mae: Self-supervised representation learning from images and video with contrastive masked autoencoders},
  author={Hernandez, Jefferson and Villegas, Ruben and Ordonez, Vicente},
  booktitle={European Conference on Computer Vision},
  pages={444--463},
  year={2024},
  organization={Springer}
}

@inproceedings{FAME,
  title={Motion-aware contrastive video representation learning via foreground-background merging},
  author={Ding, Shuangrui and Li, Maomao and Yang, Tianyu and Qian, Rui and Xu, Haohang and Chen, Qingyi and Wang, Jue and Xiong, Hongkai},
  booktitle={Proceedings of the IEEE/CVF Conference on Computer Vision and Pattern Recognition},
  pages={9716--9726},
  year={2022}
}

@inproceedings{ProViCo,
  title={Probabilistic representations for video contrastive learning},
  author={Park, Jungin and Lee, Jiyoung and Kim, Ig-Jae and Sohn, Kwanghoon},
  booktitle={Proceedings of the IEEE/CVF Conference on Computer Vision and Pattern Recognition},
  pages={14711--14721},
  year={2022}
}

@inproceedings{VCL,
  title={Static and dynamic concepts for self-supervised video representation learning},
  author={Qian, Rui and Ding, Shuangrui and Liu, Xian and Lin, Dahua},
  booktitle={European Conference on Computer Vision},
  pages={145--164},
  year={2022},
  organization={Springer}
}

@article{ST-Adapter,
  title={St-adapter: Parameter-efficient image-to-video transfer learning},
  author={Pan, Junting and Lin, Ziyi and Zhu, Xiatian and Shao, Jing and Li, Hongsheng},
  journal={Advances in Neural Information Processing Systems},
  volume={35},
  pages={26462--26477},
  year={2022}
}

@inproceedings{wu2023dropmae,
  title={Dropmae: Masked autoencoders with spatial-attention dropout for tracking tasks},
  author={Wu, Qiangqiang and Yang, Tianyu and Liu, Ziquan and Wu, Baoyuan and Shan, Ying and Chan, Antoni B},
  booktitle={Proceedings of the IEEE/CVF Conference on Computer Vision and Pattern Recognition},
  pages={14561--14571},
  year={2023}
}

@inproceedings{intrator2023endossl,
  title={Self-supervised polyp re-identification in colonoscopy},
  author={Intrator, Yotam and Aizenberg, Natalie and Livne, Amir and Rivlin, Ehud and Goldenberg, Roman},
  booktitle={International Conference on Medical Image Computing and Computer-Assisted Intervention},
  pages={590--600},
  year={2023},
  organization={Springer}
}

@article{mohammed2020endossl2,
  title={PS-DeVCEM: Pathology-sensitive deep learning model for video capsule endoscopy based on weakly labeled data},
  author={Mohammed, Ahmed and Farup, Ivar and Pedersen, Marius and Yildirim, Sule and Hovde, {\O}istein},
  journal={Computer Vision and Image Understanding},
  volume={201},
  pages={103062},
  year={2020},
  publisher={Elsevier}
}

@inproceedings{vats2021endossl3,
  title={A preliminary analysis of self-supervision for wireless capsule endoscopy},
  author={Vats, Anuja and Pedersen, Marius and Mohammed, Ahmed},
  booktitle={2021 9th European Workshop on Visual Information Processing (EUVIP)},
  pages={1--6},
  year={2021},
  organization={IEEE}
}

@inproceedings{wang2015stsep,
  title={Unsupervised learning of visual representations using videos},
  author={Wang, Xiaolong and Gupta, Abhinav},
  booktitle={Proceedings of the IEEE International Conference on Computer Vision},
  pages={2794--2802},
  year={2015}
}

@inproceedings{diba2018stsep2,
  title={Temporal 3d convnets using temporal transition layer},
  author={Diba, Ali and Fayyaz, Mohsen and Sharma, Vivek and Hossein Karami, A and Mahdi Arzani, M and Yousefzadeh, Rahman and Van Gool, Luc},
  booktitle={Proceedings of the IEEE Conference on Computer Vision and Pattern Recognition Workshops},
  pages={1117--1121},
  year={2018}
}

@inproceedings{qing2023stsep3,
  title={Disentangling spatial and temporal learning for efficient image-to-video transfer learning},
  author={Qing, Zhiwu and Zhang, Shiwei and Huang, Ziyuan and Zhang, Yingya and Gao, Changxin and Zhao, Deli and Sang, Nong},
  booktitle={Proceedings of the IEEE/CVF International Conference on Computer Vision},
  pages={13934--13944},
  year={2023}
}

@inproceedings{korbar2019stsep4,
  title={Scsampler: Sampling salient clips from video for efficient action recognition},
  author={Korbar, Bruno and Tran, Du and Torresani, Lorenzo},
  booktitle={Proceedings of the IEEE/CVF International Conference on Computer Vision},
  pages={6232--6242},
  year={2019}
}

@inproceedings{arnab2021vivitmultisep,
  title={Vivit: A video vision transformer},
  author={Arnab, Anurag and Dehghani, Mostafa and Heigold, Georg and Sun, Chen and Lu{\v{c}}i{\'c}, Mario and Schmid, Cordelia},
  booktitle={Proceedings of the IEEE/CVF International Conference on Computer Vision},
  pages={6836--6846},
  year={2021}
}

@inproceedings{liu2022multisep2,
  title={Video swin transformer},
  author={Liu, Ze and Ning, Jia and Cao, Yue and Wei, Yixuan and Zhang, Zheng and Lin, Stephen and Hu, Han},
  booktitle={Proceedings of the IEEE/CVF Conference on Computer Vision and Pattern Recognition},
  pages={3202--3211},
  year={2022}
}

@inproceedings{xiao2022hierarchicalsep,
  title={Hierarchical self-supervised representation learning for movie understanding},
  author={Xiao, Fanyi and Kundu, Kaustav and Tighe, Joseph and Modolo, Davide},
  booktitle={Proceedings of the IEEE/CVF Conference on Computer Vision and Pattern Recognition},
  pages={9727--9736},
  year={2022}
}

@inproceedings{reed2023scalesep,
  title={Scale-mae: A scale-aware masked autoencoder for multiscale geospatial representation learning},
  author={Reed, Colorado J and Gupta, Ritwik and Li, Shufan and Brockman, Sarah and Funk, Christopher and Clipp, Brian and Keutzer, Kurt and Candido, Salvatore and Uyttendaele, Matt and Darrell, Trevor},
  booktitle={Proceedings of the IEEE/CVF International Conference on Computer Vision},
  pages={4088--4099},
  year={2023}
}

@inproceedings{gu2024mamba,
  title={Mamba: Linear-time sequence modeling with selective state spaces},
  author={Gu, Albert and Dao, Tri},
  booktitle={First Conference on Language Modeling},
  year={2024}
}

@inproceedings{
gu2022S4,
title={Efficiently Modeling Long Sequences with Structured State Spaces},
author={Albert Gu and Karan Goel and Christopher Re},
booktitle={International Conference on Learning Representations},
year={2022},
url={https://openreview.net/forum?id=uYLFoz1vlAC}
}

@article{gu2020hippo,
  title={Hippo: Recurrent memory with optimal polynomial projections},
  author={Gu, Albert and Dao, Tri and Ermon, Stefano and Rudra, Atri and R{\'e}, Christopher},
  journal={Advances in Neural Information Processing Systems},
  volume={33},
  pages={1474--1487},
  year={2020}
}

@article{liu2024vmamba,
  title={Vmamba: Visual state space model},
  author={Liu, Yue and Tian, Yunjie and Zhao, Yuzhong and Yu, Hongtian and Xie, Lingxi and Wang, Yaowei and Ye, Qixiang and Jiao, Jianbin and Liu, Yunfan},
  journal={Advances in Neural Information Processing Systems},
  volume={37},
  pages={103031--103063},
  year={2024}
}

@inproceedings{
zhu2024visionmamba,
title={Vision Mamba: Efficient Visual Representation Learning with Bidirectional State Space Model},
author={Lianghui Zhu and Bencheng Liao and Qian Zhang and Xinlong Wang and Wenyu Liu and Xinggang Wang},
booktitle={Forty-first International Conference on Machine Learning},
year={2024},
url={https://openreview.net/forum?id=YbHCqn4qF4}
}

@article{jinhao2024PSTUDA,
  title={One-to-Multiple: A Progressive Style Transfer Unsupervised Domain-Adaptive Framework for Kidney Tumor Segmentation},
  author={Hu, Kai and Li, Jinhao and Zhang, Yuan and Ye, Xiongjun and Gao, Xieping},
  journal={Advances in Neural Information Processing Systems},
  volume={37},
  pages={24496--24522},
  year={2024}
}

@inproceedings{
yuzhang2025confusiondriven,
title={Confusion-Driven Self-Supervised Progressively Weighted Ensemble Learning for Non-Exemplar Class Incremental Learning},
author={Kai Hu and Zhang Yu and Yuan Zhang and Zhineng Chen and Xieping Gao},
booktitle={The Thirty-ninth Annual Conference on Neural Information Processing Systems},
year={2025},
url={https://openreview.net/forum?id=yflq8Bhjrw}
}

@article{mesejo2016Colonoscopic,
  title={Computer-aided classification of gastrointestinal lesions in regular colonoscopy},
  author={Mesejo, Pablo and Pizarro, Daniel and Abergel, Armand and Rouquette, Olivier and Beorchia, Sylvain and Poincloux, Laurent and Bartoli, Adrien},
  journal={IEEE Transactions on Medical Imaging},
  volume={35},
  number={9},
  pages={2051--2063},
  year={2016},
  publisher={IEEE}
}

@inproceedings{ma2021ldpolypvideo,
  title={LDPolypVideo benchmark: a large-scale colonoscopy video dataset of diverse polyps},
  author={Ma, Yiting and Chen, Xuejin and Cheng, Kai and Li, Yang and Sun, Bin},
  booktitle={International Conference on Medical Image Computing and Computer-Assisted Intervention},
  pages={387--396},
  year={2021},
  organization={Springer}
}

@article{borgli2020hyperkvasir,
  title={HyperKvasir, a comprehensive multi-class image and video dataset for gastrointestinal endoscopy},
  author={Borgli, Hanna and Thambawita, Vajira and Smedsrud, Pia H and Hicks, Steven and Jha, Debesh and Eskeland, Sigrun L and Randel, Kristin Ranheim and Pogorelov, Konstantin and Lux, Mathias and Nguyen, Duc Tien Dang and others},
  journal={Scientific Data},
  volume={7},
  number={1},
  pages={283},
  year={2020},
  publisher={Nature Publishing Group UK London}
}

@article{smedsrud2021kvasir,
  title={Kvasir-Capsule, a video capsule endoscopy dataset},
  author={Smedsrud, Pia H and Thambawita, Vajira and Hicks, Steven A and Gjestang, Henrik and Nedrejord, Oda Olsen and N{\ae}ss, Espen and Borgli, Hanna and Jha, Debesh and Berstad, Tor Jan Derek and Eskeland, Sigrun L and others},
  journal={Scientific Data},
  volume={8},
  number={1},
  pages={142},
  year={2021},
  publisher={Nature Publishing Group UK London}
}

@article{nwoye2022CholecTriplet,
  title={Rendezvous: Attention mechanisms for the recognition of surgical action triplets in endoscopic videos},
  author={Nwoye, Chinedu Innocent and Yu, Tong and Gonzalez, Cristians and Seeliger, Barbara and Mascagni, Pietro and Mutter, Didier and Marescaux, Jacques and Padoy, Nicolas},
  journal={Medical Image Analysis},
  volume={78},
  pages={102433},
  year={2022},
  publisher={Elsevier}
}

@inproceedings{tian2022PolypDiag,
  title={Contrastive transformer-based multiple instance learning for weakly supervised polyp frame detection},
  author={Tian, Yu and Pang, Guansong and Liu, Fengbei and Liu, Yuyuan and Wang, Chong and Chen, Yuanhong and Verjans, Johan and Carneiro, Gustavo},
  booktitle={International Conference on Medical Image Computing and Computer-Assisted Intervention},
  pages={88--98},
  year={2022},
  organization={Springer}
}

@article{bernal2015CVC-12k,
  title={WM-DOVA maps for accurate polyp highlighting in colonoscopy: Validation vs. saliency maps from physicians},
  author={Bernal, Jorge and S{\'a}nchez, F Javier and Fern{\'a}ndez-Esparrach, Gloria and Gil, Debora and Rodr{\'\i}guez, Cristina and Vilari{\~n}o, Fernando},
  journal={Computerized Medical Imaging and Graphics},
  volume={43},
  pages={99--111},
  year={2015},
  publisher={Elsevier}
}

@article{li2021KUMC,
  title={Colonoscopy polyp detection and classification: Dataset creation and comparative evaluations},
  author={Li, Kaidong and Fathan, Mohammad I and Patel, Krushi and Zhang, Tianxiao and Zhong, Cuncong and Bansal, Ajay and Rastogi, Amit and Wang, Jean S and Wang, Guanghui},
  journal={PLOS One},
  volume={16},
  number={8},
  pages={e0255809},
  year={2021},
  publisher={Public Library of Science San Francisco, CA USA}
}

@article{twinanda2016cholec80,
  title={Endonet: a deep architecture for recognition tasks on laparoscopic videos},
  author={Twinanda, Andru P and Shehata, Sherif and Mutter, Didier and Marescaux, Jacques and De Mathelin, Michel and Padoy, Nicolas},
  journal={IEEE Transactions on Medical Imaging},
  volume={36},
  number={1},
  pages={86--97},
  year={2016},
  publisher={IEEE}
}

@inproceedings{Loshchilov2017AdamW,
  title={Decoupled Weight Decay Regularization},
  author={Ilya Loshchilov and Frank Hutter},
  booktitle={International Conference on Learning Representations},
  year={2017},
  url={https://api.semanticscholar.org/CorpusID:53592270}
}

@article{paszke2019pytorch,
  title={Pytorch: An imperative style, high-performance deep learning library},
  author={Paszke, Adam and Gross, Sam and Massa, Francisco and Lerer, Adam and Bradbury, James and Chanan, Gregory and Killeen, Trevor and Lin, Zeming and Gimelshein, Natalia and Antiga, Luca and others},
  journal={Advances in Neural Information Processing Systems},
  volume={32},
  year={2019}
}

@article{CHEN2024103280Transunet,
title = {TransUNet: Rethinking the U-Net architecture design for medical image segmentation through the lens of transformers},
journal = {Medical Image Analysis},
volume = {97},
pages = {103280},
year = {2024},
issn = {1361-8415},
doi = {https://doi.org/10.1016/j.media.2024.103280},
url = {https://www.sciencedirect.com/science/article/pii/S1361841524002056},
author = {Jieneng Chen and Jieru Mei and Xianhang Li and Yongyi Lu and Qihang Yu and Qingyue Wei and Xiangde Luo and Yutong Xie and Ehsan Adeli and Yan Wang and Matthew P. Lungren and Shaoting Zhang and Lei Xing and Le Lu and Alan Yuille and Yuyin Zhou},
}

@inproceedings{wu2021STFT,
  title={Multi-frame collaboration for effective endoscopic video polyp detection via spatial-temporal feature transformation},
  author={Wu, Lingyun and Hu, Zhiqiang and Ji, Yuanfeng and Luo, Ping and Zhang, Shaoting},
  booktitle={International Conference on Medical Image Computing and Computer-Assisted Intervention},
  pages={302--312},
  year={2021},
  organization={Springer}
}

@article{jin2017sv-rcnet,
  title={SV-RCNet: workflow recognition from surgical videos using recurrent convolutional network},
  author={Jin, Yueming and Dou, Qi and Chen, Hao and Yu, Lequan and Qin, Jing and Fu, Chi-Wing and Heng, Pheng-Ann},
  journal={IEEE Transactions on Medical Imaging},
  volume={37},
  number={5},
  pages={1114--1126},
  year={2017},
  publisher={IEEE}
}
}

\clearpage
\setcounter{page}{1}
\maketitlesupplementary

\newpage
\appendix
In the supplementary material, we first summarize the endoscopic video datasets used in our study including pre-training corpora and downstream benchmarks in Section~\ref{app:A}. Next, we provide implementation details for reproducibility, covering pre-training settings, evaluation protocols, and competitors, in Section~\ref{app:B}. In Section~\ref{app:C}, we present additional experimental results including surgical phase recognition, extended ablations, and further visualizations (segmentation/detection comparisons and mask visualizations). Finally, we analyze the failure case and present future research in Section~\ref{app:D}.

\section{Dataset Details}
\label{app:A}

The datasets used in the experiment include 7 pre-training datasets and 4 downstream task datasets. Although EndoFM-LV \cite{wang2025improvingEndoFM-LV} and EndoMamba \cite{tian2025endomamba} extend the pre-training dataset, the extended datasets are not fully disclosed. Therefore, we still use publicly available datasets including Colonoscopic \cite{mesejo2016Colonoscopic}, SUN-SEG \cite{ji2022videopolyp}, LDPolypVideo \cite{ma2021ldpolypvideo}, Hyper-Kvasir \cite{borgli2020hyperkvasir}, Kvasir-Capsule \cite{smedsrud2021kvasir}, CholecTriplet \cite{nwoye2022CholecTriplet}, Renji-Hospital \cite{wang2023foundationEndoFM}, PolypDiag \cite{tian2022PolypDiag}, CVC-12k \cite{bernal2015CVC-12k}, KUMC \cite{li2021KUMC}, and Cholec80 \cite{twinanda2016cholec80} as evaluation benchmarks for all experiments.

\subsection{Pre-training Datasets}
\begin{figure*}[!b]
    \centering
    \includegraphics[width=1\linewidth]{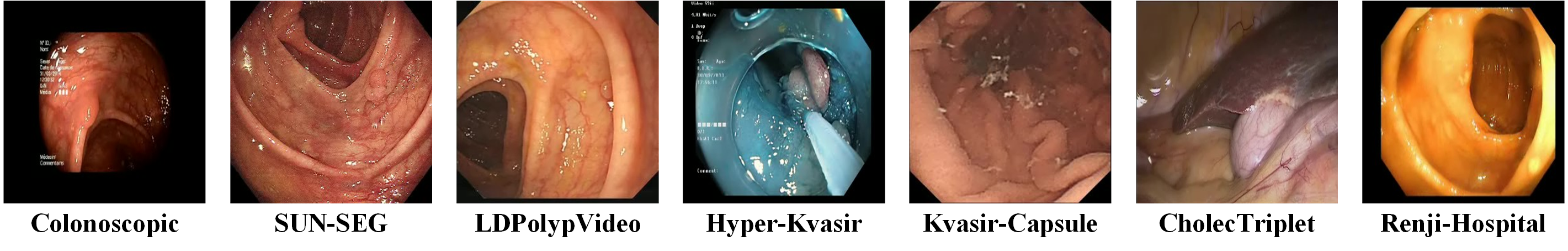}
    \caption{Example frames of the 7 pre-training datasets used in this work.}
    \label{fig:pretrain_datasets}
\end{figure*}

We leverage 7 medical video datasets spanning diagnostic endoscopy, capsule endoscopy, and laparoscopic surgery as unlabeled sources for self-supervised pre-training. In total, these datasets provide 32,896 videos with 5,024,101 frames. Unless otherwise stated, we sample 30 FPS short clips with an average duration of 5 seconds from each video. Representative frames from all datasets are shown in Fig.~\ref{fig:pretrain_datasets}, illustrating the diversity in imaging modality, anatomical site, and visual appearance.

\noindent\textbf{Colonoscopic} \cite{mesejo2016Colonoscopic} is a small-scale medical video dataset collected from routine screening colonoscopies, focusing on gastrointestinal lesions (mainly colorectal polyps) observed under both white-light and narrow-band imaging. It contains 76 short colonoscopy videos centered on individual lesions and captures realistic in-procedure appearance variations such as camera motion, specular highlights, and illumination changes.

\noindent\textbf{SUN-SEG} \cite{ji2022videopolyp} is a large-scale colonoscopy video benchmark for polyp-centric analysis, constructed from the SUN database collected at tertiary hospitals. It consists of 1,106 short video clips with 158{,}690 frames in total, covering diverse polyp sizes, morphologies, and anatomical locations under realistic screening conditions such as rapid camera motion, specular highlights, fluids, and low-contrast mucosa.

\noindent\textbf{LDPolypVideo} \cite{ma2021ldpolypvideo} is a diverse colonoscopy video dataset designed to capture real-world variability in colorectal polyps across different patients and examination settings. It contains 160 colonoscopy videos with 40{,}266 frames, where polyps exhibit wide variations in size, shape, texture, and viewing angle, together with challenging artifacts such as motion blur and occlusions.

\noindent\textbf{Hyper-Kvasir} \cite{borgli2020hyperkvasir} is a large-scale gastrointestinal endoscopy dataset collected from routine gastro- and colonoscopy examinations, covering both upper and lower GI tract with a broad spectrum of anatomical landmarks and pathological findings. We use only its video subset, which comprises 374 endoscopic recordings (about 10 hours, $\sim$0.9M frames) of real clinical procedures.

\noindent\textbf{Kvasir-Capsule} \cite{smedsrud2021kvasir} is a large-scale video capsule endoscopy dataset consisting of 117 complete small-bowel examinations acquired with wireless capsule cameras, totaling approximately 4.7M frames. The recordings capture long-range traversal of the gastrointestinal mucosa with diverse findings and image quality variations typical of capsule endoscopy.

\noindent\textbf{CholecTriplet} \cite{nwoye2022CholecTriplet} is a laparoscopic surgery video dataset built on CholecT50, containing 50 full-length cholecystectomy procedures recorded from the operative camera (about $10^5$ frames at 1 FPS). It provides long, workflow-rich sequences with frequent instrument–tissue interactions and view changes that are complementary to diagnostic endoscopy.

\noindent\textbf{Renji-Hospital} \cite{wang2023foundationEndoFM} is a large-scale clinical endoscopy video dataset collected from routine upper and lower gastrointestinal examinations at the Baoshan Branch of Renji Hospital in Shanghai, China. It comprises 16{,}494 colonoscopy clips (2,491,952 frames) and 7,653 gastroscopy clips (1,170,753 frames), covering common mucosal abnormalities such as polyps and erosions under real-world screening and diagnostic conditions.

\subsection{Downstream Datasets}
\begin{figure}
    \centering
    \includegraphics[width=1\linewidth]{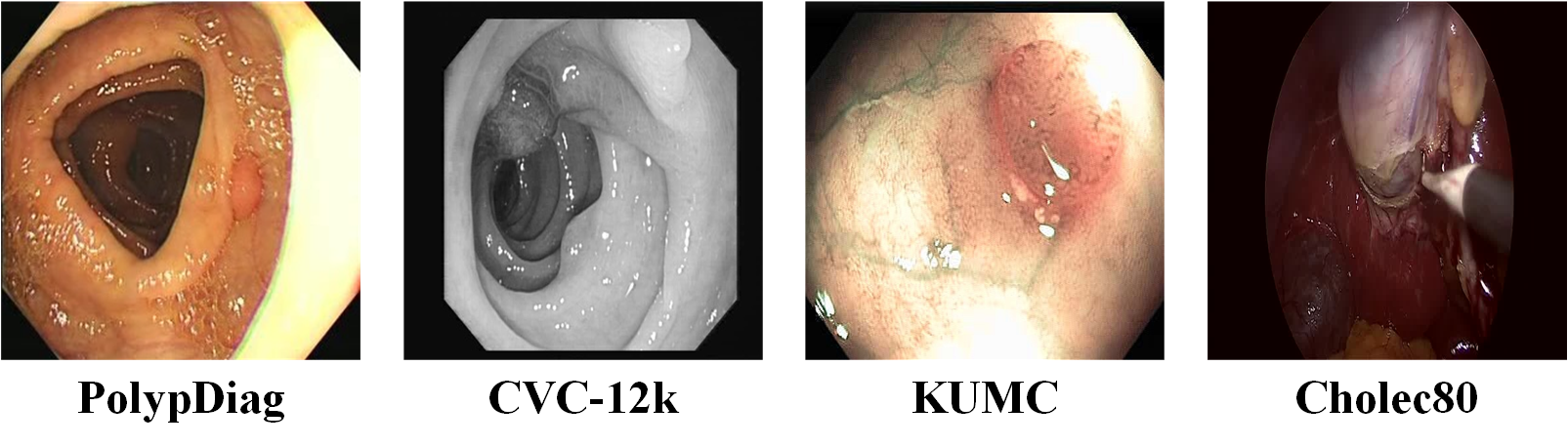}
    \caption{Example frames of the 4 downstream datasets used in this work.}
    \label{fig:downstream_datasets}
\end{figure}

We evaluate our representations on 4 labeled benchmarks covering disease diagnosis, polyp segmentation, polyp detection, and surgical phase recognition. Representative frames from these downstream datasets are shown in Fig.~\ref{fig:downstream_datasets}.

\noindent\textbf{PolypDiag} \cite{tian2022PolypDiag} is a large-scale endoscopy video dataset constructed from Hyper-Kvasir and LDPolypVideo, targeting lesion-level disease diagnosis. It contains 253 short endoscopic clips with 485{,}561 frames in total, where each clip is assigned a binary video-level label indicating the presence or absence of neoplastic lesions (polyps or early cancers). We use these video-level lesion labels to fine-tune and evaluate our model on the disease classification task.

\noindent\textbf{CVC-12k} \cite{bernal2015CVC-12k} is a colonoscopy image dataset built from 18 video sequences, comprising 11{,}954 frames of which most contain at least one colorectal polyp. Each frame is annotated with a polyp region, originally provided as approximate masks around the visible lesions. Following Endo-FM, we reorganize the annotated frames into 29 short video clips (612 labeled frames) and convert the polyp regions into pixel-wise masks. This yields a supervised benchmark for evaluating polyp segmentation performance.

\noindent\textbf{KUMC} \cite{li2021KUMC} is a colonoscopy polyp detection and classification dataset collected at the University of Kansas Medical Center. It consists of 80 video sequences curated from routine colonoscopy examinations, where frames are annotated with bounding boxes and categorical labels for individual polyps (adenomatous vs.\ hyperplastic). 
% In line with prior work, we adopt the KUMC split used in Endo-FM, which includes 53 sequences with about 19,832 labeled frames, and use the bounding-box and category annotations to assess polyp detection performance.
We use the same data partitioning method as in Endo-FM, which comprises 53 sequences containing approximately 19,832 labeled frames. We leverage the bounding-box and category annotations to evaluate the performance of polyp detection.

\noindent\textbf{Cholec80} \cite{twinanda2016cholec80} is a laparoscopic cholecystectomy dataset comprising 80 full-length surgical videos (about 40 minutes each) recorded at 25 FPS from 13 surgeons. Each frame is annotated with one of seven surgical phases, and tool-presence labels are provided at 1 FPS. We follow the standard 40/40 train–test split and use only the phase annotations to benchmark cross-domain transfer on surgical workflow recognition.

% \subsection{Metrics of Evaluation}
% To measure the quality of endoscopic video representation learning.

\section{Implementation Details}
\label{app:B}

\subsection{Pre-training Settings}
Our \textit{FPRL} is built based on EndoMamba-S \cite{tian2025endomamba}, where the patch size and embedding dimension are set to 16 and 384, respectively. For every input video, we generate 3 views with spatial size 224$\times$224 and 2 frames per view from a temporal window. The model is trained using the AdamW optimizer \cite{Loshchilov2017AdamW} with a base learning rate of 1.5e-4, a cosine learning rate schedule for 400 epochs, and a batch size of 64, with the first 40 epochs dedicated to linear warmup. For feature alignment, the pretrained VideoMamba-S \cite{li2024videomamba} serves as the teacher model. 
% in keeping with the EndoMamba setup, with the loss weight set to \(\lambda_{\mathrm{2}}\) = 0.8 and other hyper-parameters of the loss function are set as follows: \(\lambda_{\mathrm{1}}\) = 1.0, \(\lambda_{\mathrm{3}}\) =1.0, and \(\lambda_{\mathrm{pf}}\) = 20 based on preliminary experiments. 
In accordance with the EndoMamba configuration, the weight for the feature alignment loss is set to \(\lambda_{\mathrm{2}}\) = 0.8. The other hyperparameters of the loss functions are established as follows: \(\lambda_{\mathrm{1}}\) = 1.0, \(\lambda_{\mathrm{3}}\) = 1.0, and \(\lambda_{\mathrm{pf}}\) = 20, based on preliminary experiments.
% The general hyperparameters settings for our \textit{FPRL} framework during the training process are summarized in Table \ref{tab:settings}. All experiments in the process are conducted with Pytorch \cite{paszke2019pytorch} on a Linux machine equipped with 4 NVIDIA A800 GPUs.
The general hyperparameter settings for our \textit{FPRL} framework during the training process are summarized in Table \ref{tab:settings}. All experiments are conducted using PyTorch \cite{paszke2019pytorch} on a Linux machine equipped with four NVIDIA A800 GPUs.

\begin{table}[h]
\centering
\caption{Pre-training settings.}
\label{tab:settings}
\renewcommand{\arraystretch}{1.1}
\begin{tabular}{l|c}
\hline
\multicolumn{1}{c|}{\textbf{Hyperparameter}} & \textbf{Value} \\ \hline
\rowcolor{cyan!10}
\multicolumn{2}{c}{Sampling strategies} \\ \hline
temporal window length    & 50 \\
number of view frames     & 2  \\
crop size                 & \(224 \times 224\) \\
mask ratio                & 0.9 \\ \hline
\rowcolor{cyan!10}
\multicolumn{2}{c}{Optimizing settings} \\ \hline
optimizer                 & AdamW \\
learning rate schedule    & Cosine \\
weight decay              & 0.05 \\
optimizer momentum        & \(\beta_1\), \(\beta_2\) = 0.9, 0.999 \\
patch size                & 16 \\
base learning rate        & $1.5\times 10^{-4}$ \\
warm-up epochs            & 40 \\
pretraining epochs        & 400 \\
batch size of each GPU    & 64 \\
feature dimension         & 384 \\ \hline
\rowcolor{cyan!10}
\multicolumn{2}{c}{Loss functions} \\ \hline
weight of reconstruction loss & \(\lambda_{\mathrm{1}}\) = 1 \\
weight of feature align loss  & \(\lambda_{\mathrm{2}}\) = 0.8 \\
weight of contrastive loss    & \(\lambda_{\mathrm{3}}\) = 1 \\
weight of squeezing loss      & \(\lambda_{\mathrm{pf}}\) = 20 \\ \hline

\end{tabular}
\end{table}

\subsection{Evaluation Settings}
For downstream fine-tuning, we utilize the following setup on a single NVIDIA A800 GPU.
\textit{1) PolypDiag}: We sample 8 frames at a resolution of 224$\times$224 from each video as input, utilizing a pre-trained model to initialize the backbone and appending randomly initialized linear layers, and train for 20 epochs. The SGD optimizer is employed, with the learning rate set to 1e-3, momentum to 0.9, and batch size to 4.
\textit{2) CVC-12k}: A TransUNet \cite{CHEN2024103280Transunet} equipped with \textit{FPRL} as the backbone is implemented. The AdamW optimizer is used to optimize the overall parameters by setting the learning rate as 1e-4, weight decay as 5e-2 and the batch size as 1. We resize the spatial resolution to 224$\times$224 and fine-tune for 150 epochs.
\textit{3) KUMC}: We implement an STFT \cite{wu2021STFT} with our pre-trained model as backbone for generating a feature pyramid. We resize the spatial size to 640$\times$640 and train for 24k iterations. The SGD optimizer is used to optimize the overall parameters by setting the learning rate as 2.5e-3, weight decay as 1e-4 and momentum as 0.9.
\textit{4) Cholec80}: We utilize SV-RCNet \cite{jin2017sv-rcnet} for endoscopic surgical phase recognition, in which \textit{FPRL} serves as a temporal module for extracted features. The input frames are resized to 224$\times$224 and the model is trained for 25 epochs. Both \textit{FPRL} and randomly initialized modules are updated during downstream fine-tuning.
For the evaluation metrics, following the previous works, we use F1 score for PolypDiag, Dice for CVC-12k, F1 score for KUMC, and accuracy for Cholec80.

\subsection{Competitors}
We compare the proposed \textit{FPRL} with several recent SOTA approaches for endoscopy video analysis. These approaches include:
\begin{itemize}
    \item FAME \cite{FAME} utilizes foreground--background merging to alleviate background bias and enhance motion-aware video representations.
    \item ProViCo \cite{ProViCo} proposes a probabilistic video contrastive learning scheme that models clip-wise uncertainty in the latent space.
    \item VCL \cite{VCL} focuses on jointly learning static and dynamic concepts to improve video representation modeling.
    \item ST-Adapter \cite{ST-Adapter} inserts lightweight spatio-temporal adapters into frozen image backbones for parameter-efficient image-to-video transfer.
    \item VideoMAE \cite{tong2022videomae} is a masked autoencoder baseline that learns video representations by reconstructing highly masked video tubes.
    \item Endo-FM \cite{wang2023foundationEndoFM} develops a transformer-based foundation model pre-trained on large-scale endoscopic videos.
    \item DropMAE \cite{wu2023dropmae} augments masked autoencoding with spatial-attention dropout to better capture temporal correspondences for downstream tracking and segmentation tasks.
    \item VideoMAE V2 \cite{wang2023videomaev2} scales VideoMAE with dual masking and large-scale pre-training to build a general video foundation model.
    \item M$^2$CRL \cite{hu2024mmcrl} integrates multi-view masked modeling with contrastive learning, tailored for endoscopic video pre-training.
    \item VideoMamba \cite{li2024videomamba} introduces a state-space-based video backbone that models long-range spatio-temporal dynamics with linear complexity.
    \item EndoFM-LV \cite{wang2025improvingEndoFM-LV} extends Endo-FM to a minute-level pre-training framework on long endoscopy video sequences.
    \item EndoMamba \cite{tian2025endomamba} is an efficient endoscopic foundation model built on bidirectional and causal Mamba blocks under a hierarchical pre-training scheme.
\end{itemize}

\section{Additional Experimental Results}
\label{app:C}

\subsection{Surgical Phase Recognition}
To assess the generalizability of \textit{FPRL} to long-horizon reasoning, we further evaluate it on surgical phase recognition using the Cholec80 dataset. We follow the standardized protocol adopted in prior works, training on the official training videos and reporting frame-wise phase classification accuracy on the test split.
From Table~\ref{cholec80}, we can observe that the proposed \textit{FPRL} achieves 85.3~$\pm$~8.0\% accuracy, outperforming recent self-supervised and foundation-model baselines such as M$^2$CRL, VideoMamba, EndoFM-LV, and EndoMamba, which indicates that the representations learned by \textit{FPRL} transfer well to long-horizon workflow modeling.

\begin{table}[!htbp]
 \caption{ Surgical phase recognition.}
\centering
\renewcommand{\arraystretch}{1.1}
\setlength{\tabcolsep}{8pt}
\begin{adjustbox}{width=\linewidth}
\begin{tabular}{cccc}
\hline
Method & Venue   & Year & Accuracy   \\ \hline
FAME \cite{FAME}   & CVPR  & 2022  & 81.9 ± 9.2 \\
ProViCo \cite{ProViCo} & CVPR  & 2022 & 82.3 ± 8.5 \\
ST-Adapter \cite{ST-Adapter} & NeurIPS  & 2022  & 81.0 ± 8.7 \\
EndoSSL \cite{hirsch2023selfendoscopic} & MICCAI  & 2023    & 83.0 ± 8.0 \\
Endo-FM \cite{wang2023foundationEndoFM} & MICCAI & 2023 & 82.8 ± 9.1 \\
M$^2$CRL \cite{hu2024mmcrl} & NeurIPS & 2024   & 83.5 ± 8.8 \\ 
VideoMamba \cite{li2024videomamba} & ECCV & 2024 & 84.1 ± 8.9 \\
EndoFM-LV \cite{wang2025improvingEndoFM-LV} & JBHI  & 2025 & 85.1 ± 7.9 \\
EndoMamba \cite{tian2025endomamba} & MICCAI  & 2025 & 84.4 ± 8.4 \\
\hline
FPRL  & Ours & - & \textbf{85.3 ± 8.0} \\
\hline
\end{tabular}
\end{adjustbox}
\label{cholec80}
\end{table}

\subsection{Additional Ablations}
\noindent\textbf{Ablation on Different Architectures.}
% As shown in Table~\ref{tab:arch_ablation}, scaling the backbone from EndoMamba-T to EndoMamba-S consistently improves all three downstream tasks, especially detection (+8.1\%), while further scaling to EndoMamba-M mainly benefits segmentation but slightly degrades classification and detection, indicating that the small variant offers a better capacity–performance trade-off.
As shown in Table~\ref{tab:arch_ablation}, scaling the backbone from EndoMamba-T to EndoMamba-S consistently improves performance across all three downstream tasks, particularly in detection, which shows an improvement of +8.1\%. However, further scaling to EndoMamba-M primarily benefits segmentation while slightly degrading classification and detection performance. This suggests that the small variant provides a more favorable balance between capacity and performance.

\begin{table}[!tbp]
\caption{Ablation on model architecture variants.}
\centering
\footnotesize
\renewcommand{\arraystretch}{1.25}
\setlength{\tabcolsep}{5.5pt}
\begin{adjustbox}{width=\linewidth}
\begin{tabular}{c|ccc}
\hline
\multirow{2}{*}{Model Variant} & \multicolumn{3}{c}{Performance (\%)} \\
\cline{2-4}
& Cla. & Seg. & Det. \\ \hline
EndoMamba-T  & 92.2 ± 0.4  & 83.8 ± 0.3  & 81.7 ± 0.9  \\
EndoMamba-S  & \textbf{95.2 ± 0.3} & 86.1 ± 0.1 & \textbf{89.8 ± 0.1} \\
EndoMamba-M  & 92.3 ± 0.1  & \textbf{87.4 ± 0.5}  & 89.5 ± 0.2  \\ \hline
\end{tabular}
\end{adjustbox}
\label{tab:arch_ablation}
\end{table}

\noindent\textbf{Ablation on Auxiliary Module.}
% Table~\ref{tab:layers} analyzes the impact of decoder depth and the number of cross-view masked feature completion (CVMFC) blocks. A shallow 4-layer decoder with a single CVMFC block already achieves the best overall performance, whereas deeper decoders and multiple CVMFC blocks provide marginal or even negative gains, suggesting that overly deep temporal decoding tends to over-smooth features and hamper dense prediction.
Table~\ref{tab:layers} analyzes the impact of decoder depth and the number of cross-view masked feature completion (CVMFC) blocks. Notably, a shallow 4-layer decoder equipped with a single CVMFC block achieves optimal overall performance. In contrast, deeper decoders and multiple CVMFC blocks yield only marginal gains or have adverse effects. This observation suggests that excessive temporal decoding may lead to over-smoothing of features, ultimately hindering dense prediction capabilities.

\begin{table}[!tbp]
\caption{Ablation on Decoder depth and CVMFC blocks.}
\centering
\renewcommand{\arraystretch}{1.2}
\setlength{\tabcolsep}{6pt}
\begin{adjustbox}{width=\linewidth}
\begin{tabular}{c|c|ccc}
\hline
\multicolumn{2}{c|}{Stacked Layers} & \multicolumn{3}{c}{Performance (\%)} \\
\cline{1-2}\cline{3-5}
Decoder & CVMFC & Cla. & Seg. & Det. \\ \hline
\multirow{3}{*}{4} & 1 & 95.2 ± 0.3  & \textbf{86.1 ± 0.1} & \textbf{89.8 ± 0.1} \\
 & 2 & 94.4 ± 0.5  & 85.3 ± 0.8 & 85.2 ± 1.0 \\
 & 3 & 92.9 ± 0.4  & 86.1 ± 0.7 & 87.4 ± 0.9 \\ \hline
\multirow{3}{*}{8} & 1 & 93.8 ± 0.4  & 85.4 ± 0.5 & 84.6 ± 0.4 \\
 & 2 & \textbf{96.0 ± 0.7}  & 85.5 ± 0.8 & 87.7 ± 0.8 \\
 & 3 & 89.9 ± 0.6  & 84.9 ± 0.2 & 86.7 ± 0.9 \\ \hline
\end{tabular}
\end{adjustbox}
\label{tab:layers}
\end{table}

\noindent\textbf{Ablation on Loss Formulations.}
% As reported in Table~\ref{tab:loss_ablation}, the cosine similarity combined with an $\ell_2$ regression term clearly outperforms all other loss formulations on classification, segmentation, and detection. Replacing either cosine with cross-entropy or $\ell_2$ with $\ell_1$ leads to a noticeable drop, confirming that regressing continuous teacher features with a cosine+$\ell_2$ objective is better suited than discrete classification-based alignment.
As reported in Table~\ref{tab:loss_ablation}, the combination of cosine similarity with an $\ell_2$ regression term significantly outperforms all other loss formulations across classification, segmentation, and detection tasks. Substituting either cosine similarity with cross-entropy or $\ell_2$ with $\ell_1$ results in a marked decline in performance, thereby confirming that regressing continuous teacher features using a cosine + $\ell_2$ objective is more effective than alignment based on discrete classification.

\noindent\textbf{Ablation on the Number of Sampled Frames.}
% Table~\ref{tab:sample_frames} studies the effect of the number of sampled frames. Using two sparsely sampled frames yields the best overall performance and improves over single-frame training, whereas adopting three frames consistently harms all tasks, likely because additional views introduce redundant non-semantic motion in endoscopic videos.
Table~\ref{tab:sample_frames} investigates the effects of the number of sampled frames. Utilizing two sparsely sampled frames yields optimal overall performance and surpasses single-frame training. However, incorporating three frames consistently detracts from performance across all tasks. This decline is likely due to the introduction of redundant non-semantic motion present in endoscopic videos when additional views are included.

\begin{table}
\caption{Ablation on loss formulations.}
\centering
\renewcommand{\arraystretch}{1.2}
\setlength{\tabcolsep}{6pt}
\begin{adjustbox}{width=\linewidth}
\begin{tabular}{c|c|ccc}
\hline
\multicolumn{2}{c|}{Loss Formulations} & \multicolumn{3}{c}{Performance (\%)} \\
\cline{1-2}\cline{3-5}
Similarity & Regression & Cla. & Seg. & Det. \\ \hline
\multirow{2}{*}{Cosine} & $\ell_1$ & 92.3 ± 0.5 & 85.0 ± 0.8 & 87.3 ± 1.0 \\
 & $\ell_2$ & \textbf{95.2 ± 0.3} & \textbf{86.1 ± 0.1} & \textbf{89.8 ± 0.1} \\ \hline
\multirow{2}{*}{Cross-Entropy} & $\ell_1$ & 91.1 ± 0.5 & 80.3 ± 0.7 & 85.9 ± 0.9 \\
 & $\ell_2$ & 92.9 ± 0.7 & 82.6 ± 0.8 & 87.5 ± 0.8 \\ \hline
\end{tabular}
\end{adjustbox}
\label{tab:loss_ablation}
\end{table}

\begin{table}
\caption{Ablation on the number of sampled frames.}
\centering
\footnotesize
\renewcommand{\arraystretch}{1.2}
\setlength{\tabcolsep}{6pt}
\label{tab:sample_frames}
\begin{adjustbox}{width=\linewidth}
\begin{tabular}{c|ccc}
\hline
\multirow{2}{*}{Number of Frames} & \multicolumn{3}{c}{Performance (\%)} \\
\cline{2-4}
& Cla. & Seg. & Det. \\ \hline
1 & 94.4 $\pm$ 1.6 & 86.0 $\pm$ 0.9 & 88.6 $\pm$ 1.1 \\
% \rowcolor{pink!28}
2 & \textbf{95.2 $\pm$ 0.3}  & \textbf{86.1 $\pm$ 0.1}  & \textbf{89.8 $\pm$ 0.1} \\
3 & 93.0 $\pm$ 0.5 & 85.2 $\pm$ 0.4 & 85.4 $\pm$ 0.1 \\ \hline
\end{tabular}
\end{adjustbox}
\end{table}

\subsection{Additional Visualization Results}
\begin{figure*}[t]
    \centering
    \includegraphics[width=0.9\linewidth]{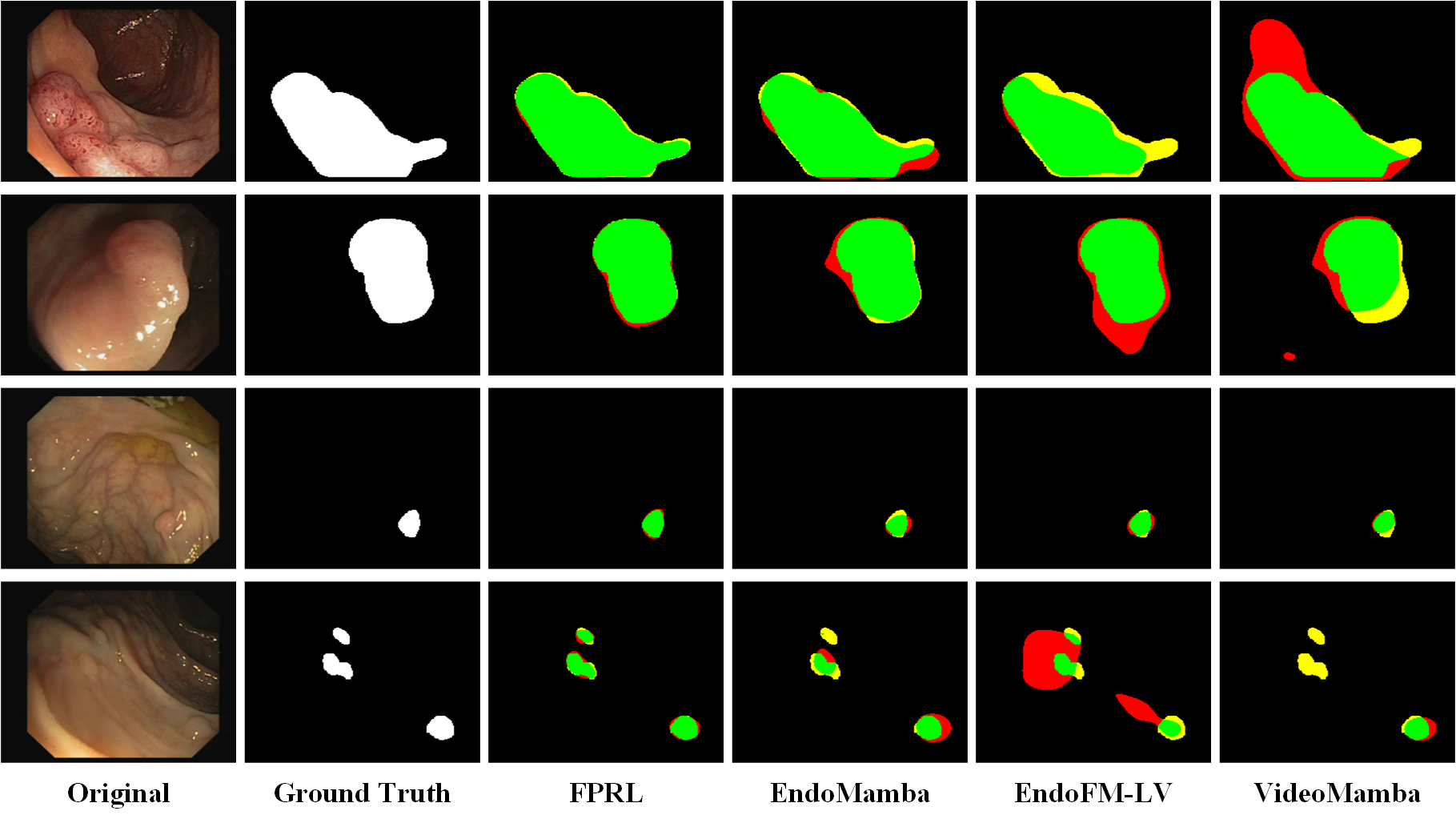}
    \caption{Qualitative results for segmentation task on the CVC-12k dataset, where green, red, and yellow regions represent the
true positive, false positive, and false negative, respectively.}
    \label{fig:seg_visual}
\end{figure*}

% Comparison chart of segmentation and detection effect, mask visualization.
\noindent\textbf{Qualitative Results for Segmentation.}
Fig.~\ref{fig:seg_visual} shows a visual comparison of segmentation results between our method and other self-supervised pre-training approaches on the CVC-12k dataset.
% Polyps often exhibit blurry boundaries and considerable shape variations, posing significant challenges to accurate segmentation. Despite these difficulties, our approach consistently produces superior results compared to other state-of-the-art self-supervised methods. 
Polyps frequently present with indistinct boundaries and considerable variations in shape, which pose significant challenges for accurate segmentation. Nevertheless, our approach consistently yields superior results when compared to other leading self-supervised methods.
% In particular, while other methods frequently misclassify or overlook certain lesion regions, as shown in the first and second rows for larger polyps, our method effectively discriminates polyps from normal tissues. 
In particular, as shown in the first and second rows for larger polyps, while other methods often misclassify or overlook certain lesion regions, our approach effectively distinguishes polyps from normal tissues.  
% Moreover, although all methods achieve comparable performance in segmenting isolated small polyps, only our method successfully delineates all polyps when multiple small instances appear together within a single frame.
Moreover, although all methods exhibit comparable performance in segmenting isolated small polyps, only our method successfully delineates all polyps when multiple small instances are present together within a single frame.

\begin{figure*}[!htbp]
    \centering
    \includegraphics[width=0.9\linewidth]{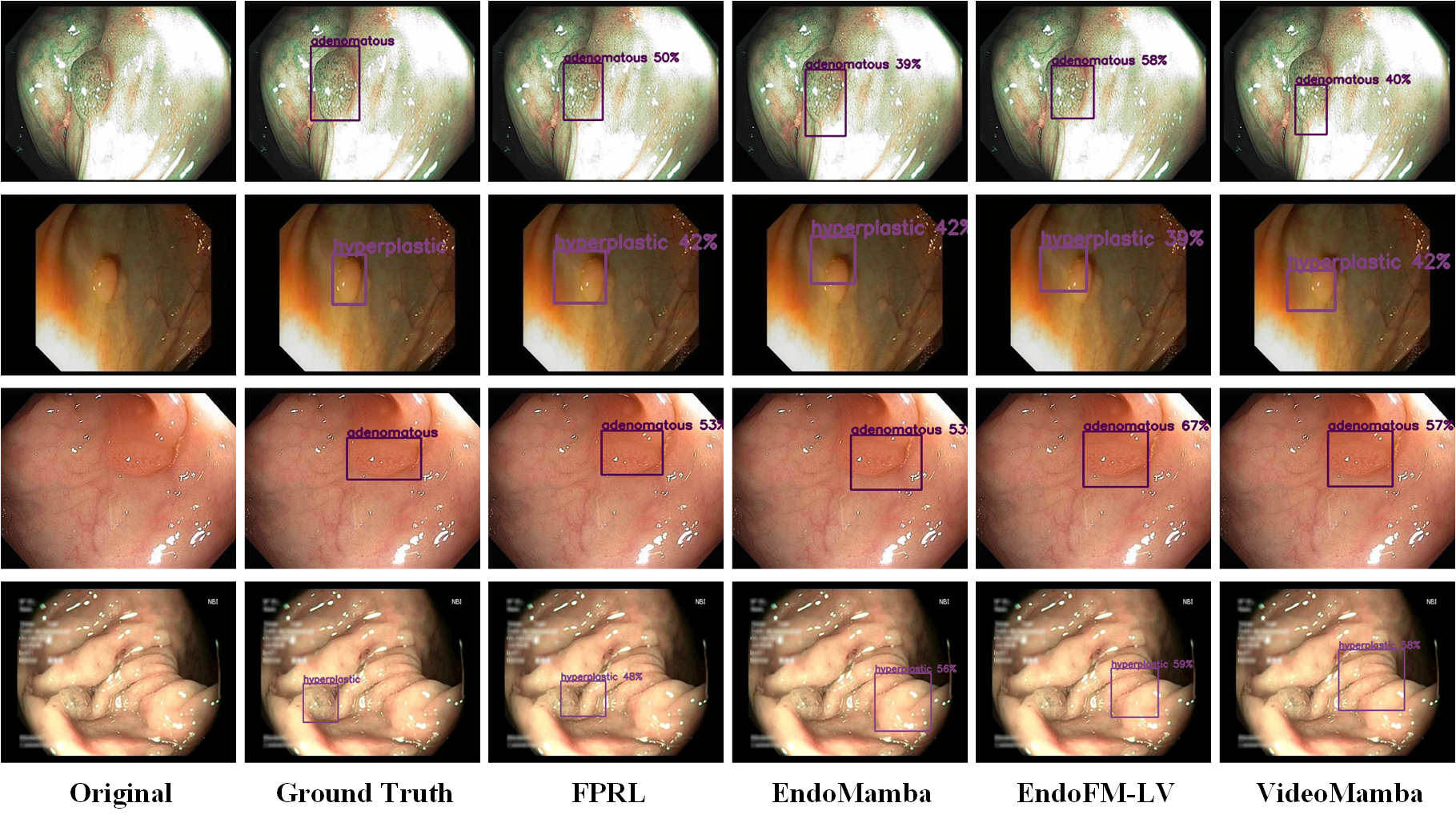}
    \caption{Qualitative results for detection task on the KUMC dataset.}
    \label{fig:det_visual}
\end{figure*}

\noindent\textbf{Qualitative Results for Detection.}
% Fig. \ref{fig:det_visual} illustrates detection results from our method and competing self-supervised pre-training techniques on the KUMC dataset. Our method demonstrates strong performance in both boundary recognition and localization of small polyps. While all methods can roughly identify the actual lesion area in high-contrast scenarios (as seen in the first and second rows), others tend to include extraneous normal tissue, whereas our method achieves higher localization precision. Furthermore, even under challenging conditions, such as low contrast, blurry polyp boundaries, or complex backgrounds, our approach remains robust in accurately identifying polyp locations. This capability stems from the effective representation learning of salient lesion regions through our proposed \textit{Static Semantic Focus} mechanism.
Fig.~\ref{fig:det_visual} presents the detection results obtained from our method and other competing self-supervised pre-training models applied to the KUMC dataset. 
Our approach exhibits superior performance in both boundary recognition and localization of small polyps. 
Although all methods can generally identify the actual lesion area in high-contrast scenarios (as shown in the first and second rows), other methods tend to incorporate extraneous normal tissue, whereas our approach achieves a higher level of localization precision. 
Moreover, even under challenging conditions, such as low contrast, blurred polyp boundaries, or complex backgrounds, our approach remains robust in accurately identifying polyp locations. 
This capability is attributed to the effective representation learning of salient lesion regions facilitated by our proposed \textit{Static Semantic Focus} mechanism.

\begin{figure*}[!htbp]
    \centering
    \includegraphics[width=0.9\linewidth]{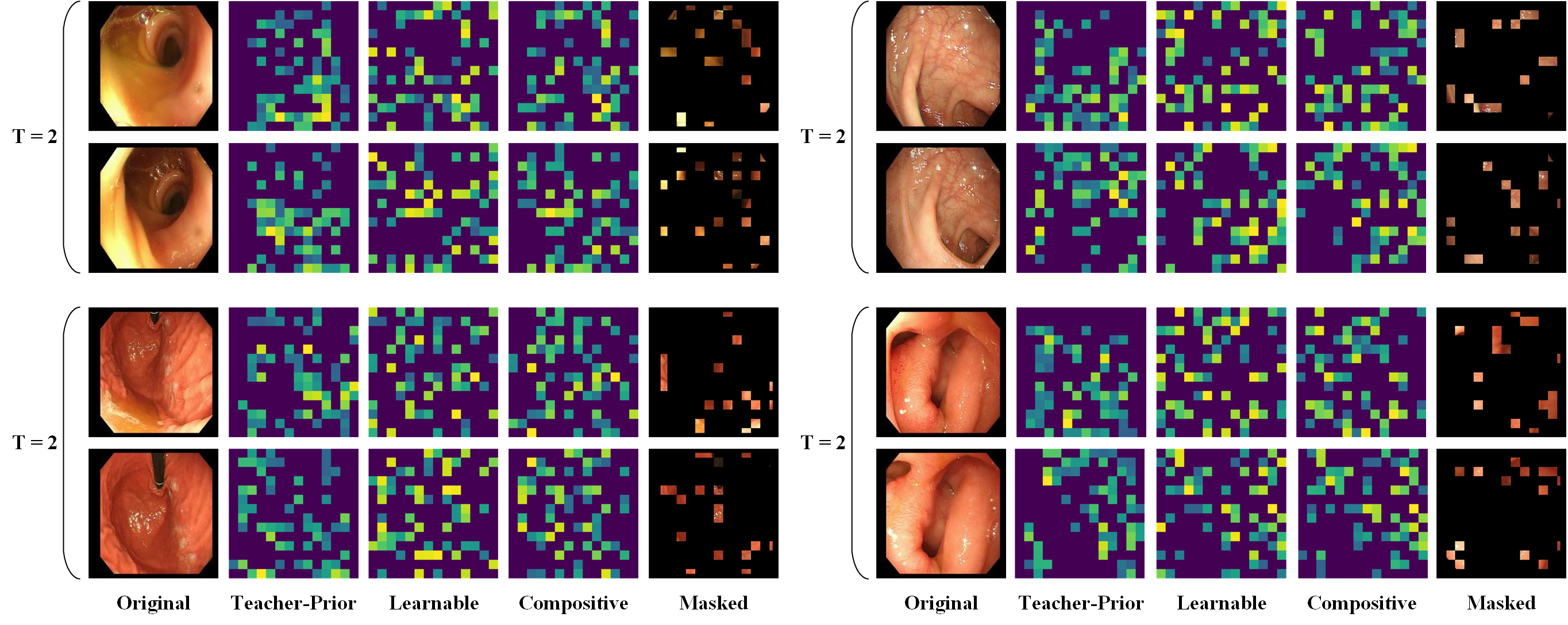}
    \caption{Illustration of our teacher-prior adaptive masking strategy. We visualize the feature heatmaps from both the teacher and the lightweight network (columns 2 \& 3). The important regions (column 4) are then selected through a weighted aggregation of these heatmaps, where visible patches are sampled via a Top-K strategy.}
    \label{fig:mask_visual}
\end{figure*}

\noindent\textbf{Mask Generation Process.}
Figure~\ref{fig:mask_visual} illustrates our teacher-prior adaptive masking strategy. We generate a saliency map from the frozen teacher network, capturing lesion-aware global priors, while a lightweight network produces a complementary map emphasizing view-specific cues such as local contrast. These are fused into a unified importance map, where we employ a Top-K sampling strategy to select the most informative regions, retaining only the corresponding patches while masking out the remainder. This process suppresses irrelevant regions like background and specular reflections, directing model focus to clinically meaningful structures. The resulting masks (shown in the last column) concentrate on polyp interiors and boundaries, promoting more focused reconstruction and stable pre-training.

\section{Failure Case Analysis}
\label{app:D}
\subsection{Failure Case}
\begin{figure*}[!htbp]
    \centering
    \includegraphics[width=1\linewidth]{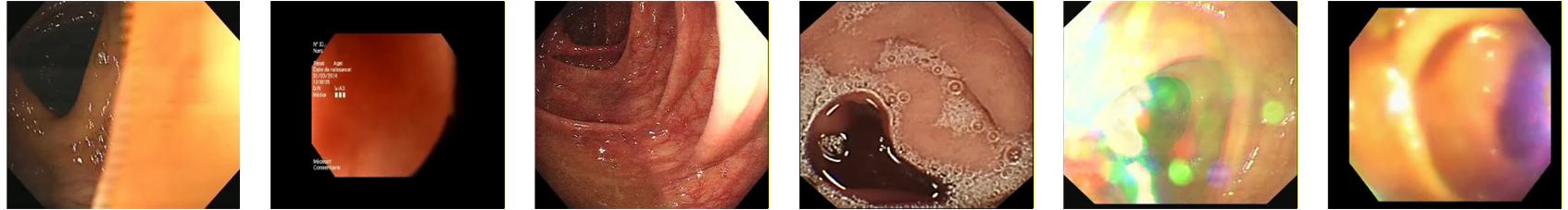}
    \caption{Examples of low-quality sampling frames.}
    \label{fig:low_quality}
\end{figure*}

As reported in Table \ref{tab:sample_frames}, the single-frame setting consistently underperforms the two-frame configuration, which appears at odds with our design objective of suppressing dynamic redundancy and non-semantic motion by keeping each view as short as possible. We argue that this discrepancy mainly stems from the imperfect quality of real endoscopic videos. In practice, many sequences contain frames that are heavily affected by motion blur, abrupt camera shake, illumination flicker, specular highlights, or transient occlusions from tools and fluids (see Fig. \ref{fig:low_quality}). Under the single-frame regime, such degradations directly contaminate the only available observation in a view, thereby corrupting the supervision signal for reconstruction and making the learned representations highly sensitive to occasional low-quality frames.

\subsection{Future Research}
From a broader perspective, the above failure case indicates that the gap between the single-frame and two-frame settings is largely driven by data quality issues rather than by the sparse-view design itself. This suggests three main directions for further improvement. 
First, the frame sampling strategy could be made more robust to low-quality observations, so that views are less likely to be dominated by severe blur, illumination fluctuation, or transient occlusions. 
Second, the pre-training framework could be refined to leverage sparse multi-frame information in a more systematic manner, utilizing additional frames primarily to enhance robustness while still guiding the model to focus on semantically meaningful dynamics.
Finally, dataset curation and augmentation strategies that explicitly account for typical endoscopic artifacts may further narrow the performance gap observed in Table~\ref{tab:sample_frames}. We leave a systematic exploration of these directions for future research.

% \clearpage

\end{document}